%% file: main.tex
\pdfoutput=1

\documentclass[11pt]{article}

\usepackage[final]{acl}

\usepackage{times}
\usepackage{latexsym}
\usepackage[T1]{fontenc}
\usepackage[utf8]{inputenc}
\usepackage{microtype}
\usepackage{inconsolata}

\usepackage{amsmath}
\usepackage{amssymb}
\usepackage{booktabs}
\usepackage{graphicx}
\usepackage{algorithm}
\usepackage{algorithmic}
\usepackage{multirow}
\usepackage{xcolor}
\usepackage{subcaption}
\usepackage{stfloats}  
\usepackage{float}  
\usepackage{enumitem}  
\usepackage[most]{tcolorbox}  
\usepackage{tikz}  

\newcommand{\method}{PAPO}
\newcommand{\aout}[1][]{A_{\text{out}\ifx\relax#1\relax\else,#1\fi}}
\newcommand{\aproc}[1][]{A_{\text{proc}\ifx\relax#1\relax\else,#1\fi}}
\newcommand{\atotal}[1][]{A_{\text{total}\ifx\relax#1\relax\else,#1\fi}}
\newcommand{\good}[1]{{\scriptsize\color{green!60!black}\textbf{#1}}}
\newcommand{\bad}[1]{{\scriptsize\color{red!70!black}\textbf{#1}}}
\newcommand{\msd}[1]{{\scriptsize$_{\pm#1}$}}
\title{PAPO: Stabilizing Rubric Integration Training via Decoupled
Advantage Normalization}

\author{
  \mdseries
  Zelin Tan$^{1,2}$ \quad
  Zhouliang Yu$^{3}$ \quad
  Bohan Lin$^{1}$ \quad
  Zijie Geng$^{1}$ \quad
  Hejia Geng$^{4}$ \quad
  Yudong Zhang$^{1}$ \\
  Mulei Zhang$^{5}$ \quad
  Yang Chen$^{2}$ \quad
  Shuyue Hu$^{2}$ \quad
  Zhenfei Yin$^{4\dagger}$ \quad
  Chen Zhang$^{2\dagger}$ \quad
  Lei Bai$^{2}$ \\[6pt]
  $^{1}$University of Science and Technology of China \quad
  $^{2}$Shanghai Artificial Intelligence Laboratory \\
  $^{3}$The Chinese University of Hong Kong \quad
  $^{4}$University of Oxford \quad
  $^{5}$Wuhan University \\[4pt]
  {\small $^{\dagger}$Corresponding authors} \\[6pt]
  \href{https://github.com/tanzelin430/PAPO}{%
    \raisebox{-1pt}{\begin{tikzpicture}[scale=0.13]
      \fill[black] (0,0) circle (1);
      \fill[white] (0,0.15) circle (0.42);
      \fill[white] (-0.32,0) rectangle (0.32,-0.55);
      \fill[white] (-0.15,-0.55) rectangle (0.15,-0.35);
      \fill[white] (-0.55,-0.45) rectangle (-0.32,-0.1);
      \fill[white] (0.32,-0.45) rectangle (0.55,-0.1);
      \fill[white] (-0.25,0.45) -- (-0.5,0.9) -- (-0.05,0.52) -- cycle;
      \fill[white] (0.25,0.45) -- (0.5,0.9) -- (0.05,0.52) -- cycle;
    \end{tikzpicture}}%
    \,\small\texttt{Code}}
  \quad
  \href{https://huggingface.co/datasets/Artemis0430/NuminaMath-20k-Stratified}{%
    \raisebox{-1pt}{\begin{tikzpicture}[scale=0.13]
      \definecolor{hfyellow}{HTML}{FFD21E}
      \fill[hfyellow] (0,0) circle (1);
      \fill[black] (-0.35,0.2) circle (0.13);
      \fill[black] (0.35,0.2) circle (0.13);
      \draw[black, line width=1.2pt, line cap=round] (-0.4,-0.25) .. controls (-0.1,-0.55) and (0.1,-0.55) .. (0.4,-0.25);
    \end{tikzpicture}}%
    \,\small\texttt{Dataset}}
}

\hypersetup{
  pdftitle={PAPO: Stabilizing Rubric Integration Training via Decoupled Advantage Normalization},
  pdfauthor={Zelin Tan, Zhouliang Yu, Bohan Lin, Zijie Geng, Hejia Geng, Yudong Zhang, Mulei Zhang, Yang Chen, Shuyue Hu, Zhenfei Yin, Chen Zhang, Lei Bai}
}

\begin{document}
\maketitle

\begin{abstract}
\input{sections/abstract}
\end{abstract}

\input{sections/introduction}
\input{sections/background}
\input{sections/problem}
\input{sections/method}
\input{sections/experiments}
\input{sections/ablation}
\input{sections/validation}
\input{sections/related}

\input{sections/conclusion}
\input{sections/limitations}


\bibliography{custom}

\appendix
\input{sections/appendix}

\end{document}

%% file: sections/abstract.tex
We propose Process-Aware Policy Optimization (\method{}), a method that integrates process-level evaluation into Group Relative Policy Optimization (GRPO) through \textbf{decoupled advantage normalization}, to address two limitations of existing reward designs. Outcome reward models (ORM) evaluate only final-answer correctness, treating all correct responses identically, and gradually lose the advantage signal as groups become uniformly correct. Rubric-based process reward models (PRM) offer richer supervision, differentiating quality among correct responses and restoring advantage diversity, but using their scores as rewards triggers verbosity-driven reward hacking and accuracy collapse. \method{} addresses both by composing the advantage from an outcome advantage $\aout$, derived from outcome reward $r^{\text{out}}$ over all responses, and a process advantage $\aproc$, derived from process reward $r^{\text{proc}}$ exclusively over the correct subset. $\aout$ anchors correctness, and correct subset normalization on $\aproc$ rules out verbosity-driven hacking while sustaining the gradient on all-correct groups where ORM exhausts. Across five model configurations and six benchmarks, \method{} improves both aggregate scores in every configuration and wins 28 of 30 individual comparisons (reaching 51.3\% vs.\ 46.3\% on OlympiadBench for Qwen2.5-7B), with analogous gains when composed with mainstream GRPO variants.

%% file: sections/introduction.tex
\section{Introduction}
\label{sec:intro}

As large language models (LLMs) \citep{yang2024qwen2,gemmateam2025gemma3technicalreport,touvron2023llamaopenefficientfoundation} continue to advance, reasoning has become a key capability for accomplishing complex tasks \citep{guo2025deepseek}. Reinforcement learning (RL) plays an essential role in improving the reasoning capabilities of LLMs, especially in mathematics \citep{deepseek-math-v2}. Among RL methods, Group Relative Policy Optimization (GRPO) \citep{shao2024deepseekmath} with outcome reward models (ORM) has emerged as a core approach in rule-based verifiable reinforcement learning for its simplicity. It eliminates the learned critic network used in PPO \citep{schulman2017proximal}, instead computing advantages through group-level reward normalization. However, GRPO with ORM evaluates only final-answer correctness, providing no signal on reasoning process quality, which leads to two critical issues.

\begin{figure*}[t]
\centering
\begin{subfigure}[b]{0.56\textwidth}
    \centering
    \includegraphics[width=\textwidth]{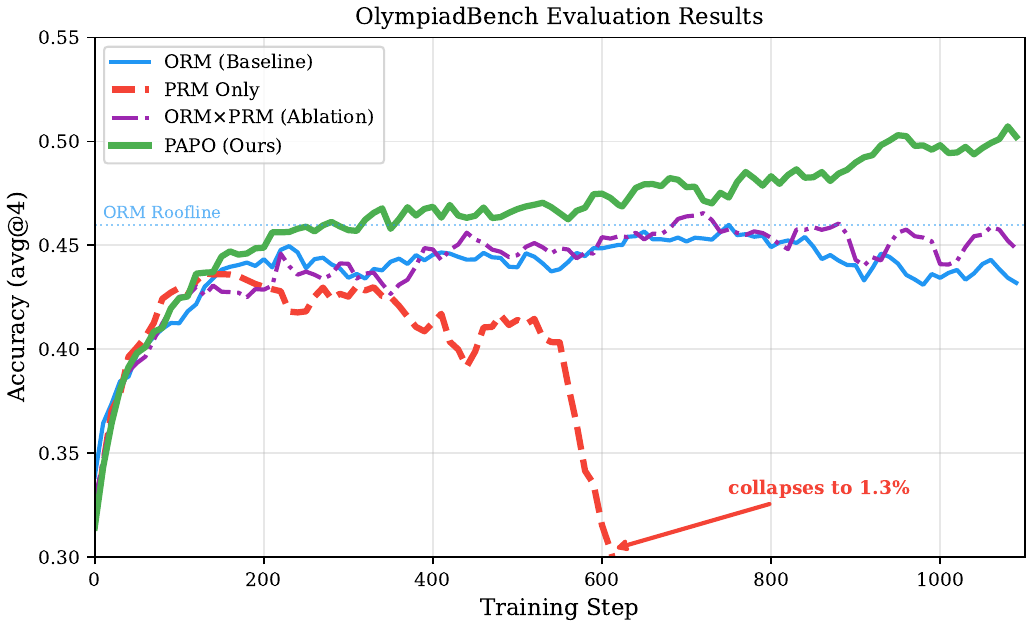}
    \caption{OlympiadBench evaluation results.}
    \label{fig:hero-a}
\end{subfigure}
\hfill
\begin{subfigure}[b]{0.42\textwidth}
    \centering
    \includegraphics[width=\textwidth]{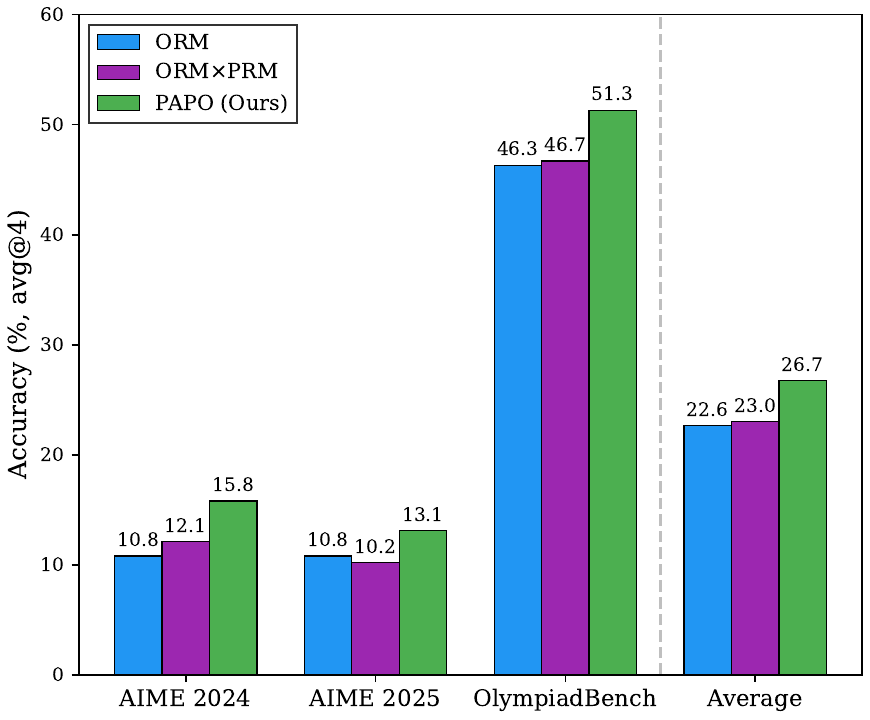}
    \caption{Accuracy on competition math.}
    \label{fig:hero-b}
\end{subfigure}
\caption{\textbf{(a)} ORM (blue) plateaus and declines after step 750 due to signal exhaustion; PRM (red, dashed) collapses via reward hacking; ORM$\times$PRM (purple, dash-dotted) tracks ORM closely without exceeding it; \method{} (green) continues improving throughout training, reaching 51.3\%. \textbf{(b)} Comparison across AIME 2024/2025, OlympiadBench, and their average. Naive multiplicative combination (ORM$\times$PRM) barely improves over ORM, while \method{}'s decoupled normalization yields substantial gains on all benchmarks.}
\label{fig:hero}
\end{figure*}

First, all correct responses are assigned identical advantage irrespective of reasoning quality, providing the same credit for lucky guesses as for rigorous step-by-step derivations. Second, as the model improves during training, an increasing proportion of response groups become uniformly correct, yielding zero advantage and zero gradient. This failure, which we call signal exhaustion, causes performance to plateau and decline (Figure~\ref{fig:hero}a).

Rubric-based process reward models (PRM) \citep{yuan2025miraclesteps,Sheng2026ReinforcingCR} offer a natural remedy by providing richer supervision that can both differentiate reasoning quality among correct responses and restore advantage diversity, addressing the two ORM failures above. However, directly using their scores as GRPO rewards leads to reward hacking \citep{skalse2022defining,mahmoud2026rewardhackingrubricbasedreinforcement}, where models inflate PRM scores by generating increasingly verbose responses while accuracy collapses. The naive multiplicative combination $r = r^{\text{out}} \times r^{\text{proc}}$ also fails to surpass ORM, since a single GRPO normalization suppresses the process signal in mixed-correctness groups, as shown in Figure~\ref{fig:hero}b.

To solve this problem, we propose \textbf{Process-Aware Policy Optimization (\method{})}, which integrates rubric-based process evaluation into GRPO at the advantage layer through \textbf{decoupled advantage normalization}. \method{} composes the advantage from two independently normalized components.
\begin{itemize}
    \item The \textbf{outcome advantage} $\aout$ is normalized from the binary outcome reward $r^{\text{out}}$ over all responses, anchoring the training signal on answer correctness.
    \item The \textbf{process advantage} $\aproc$ is normalized from the rubric-based process reward $r^{\text{proc}}$ among correct responses, differentiating reasoning quality within the correct subset.
\end{itemize}
The decoupled design offers two structural merits. Restricting $\aproc$ to the correct subset prevents incorrect responses from accruing positive process advantage, ruling out the verbosity-driven hacking pathway. At the same time, $\aproc$ remains active in all-correct groups where $\aout$ vanishes, sustaining the gradient signal precisely when ORM exhausts.

As shown in Figure~\ref{fig:hero}a, \method{} continues improving throughout training, reaching a peak of 51.3\% on OlympiadBench \citep{he2024olympiadbench} while ORM peaks at 46.3\% and declines, and Figure~\ref{fig:hero}b shows that the improvement generalizes to other math competition benchmarks like AIME \citep{balunovic_srimatharena_2025}.

To summarize, our contributions are as follows:
\begin{enumerate}
    \item We identify a dilemma in GRPO reward design: binary ORM lacks process supervision and suffers from signal exhaustion, yet naively using PRM scores causes reward hacking and training collapse.
    \item We propose \method{}, which composes the advantage from independently normalized outcome and rubric-based process components with correct subset normalization, addressing this dilemma.
    \item Across five model configurations and six benchmarks, \method{} improves both aggregate scores in every configuration and wins 28 of 30 individual comparisons, with analogous gains when composed with mainstream GRPO variants.
\end{enumerate}

%% file: sections/background.tex
\section{Background}
\label{sec:background}

\subsection{Group Relative Policy Optimization}
\label{sec:bg-grpo}

GRPO \citep{shao2024deepseekmath} replaces PPO's \citep{schulman2017proximal} learned critic with group-level reward normalization. For each prompt $x$, the model samples $G$ responses $\{o_1, \ldots, o_G\}$ from $\pi_{\theta_{\text{old}}}$, giving the objective
\begin{equation}
\label{eq:grpo-loss}
\resizebox{\columnwidth}{!}{$
\begin{aligned}
\mathcal{L}_{\text{GRPO}}=
\frac{1}{G}\sum_{i=1}^{G}\frac{1}{|o_i|}\sum_{t=1}^{|o_i|}
\Bigl\{
  \min\Bigl[
  \rho(\theta)\,\hat{A}_{i,t},\;
  \operatorname{clip}\!\Bigl(\rho(\theta), 1\!-\!\varepsilon,\, 1\!+\!\varepsilon\Bigr)\hat{A}_{i,t}
  \Bigr]
  -\beta\,\mathrm{D}_{\mathrm{KL}}
\Bigr\},
\end{aligned}
$}
\end{equation}
where $\rho(\theta) = \pi_\theta(o_{i,t} | x, o_{i,<t}) / \pi_{\theta_{\text{old}}}(o_{i,t} | x, o_{i,<t})$ is the importance sampling ratio. Each response $o_i$ receives a scalar reward $r_i$, and the advantage is
\begin{equation}
\label{eq:grpo-adv}
\hat{A}_{i,t} = \frac{r_i - \mathrm{mean}(\mathbf{r})}{\mathrm{std}(\mathbf{r})}.
\end{equation}

\subsection{Reward Models for Mathematical Reasoning}
\label{sec:bg-rm}

\paragraph{Outcome Reward Models (ORM).}
For mathematical reasoning, the ORM is a rule-based answer checker \citep{hendrycks2021measuring} that returns $r^{\text{out}} = \mathbf{1}[\hat{a} = a^*]$, where $\hat{a}$ is the predicted answer and $a^*$ the ground truth. The signal is reliable but carries no information about reasoning quality.

\paragraph{Process Reward Models (PRM).}
PRMs evaluate the reasoning process itself. We focus on \emph{rubric-based PRMs} \citep{yuan2025miraclesteps,Sheng2026ReinforcingCR}, in which an LLM-as-Judge \citep{zheng2023judgingllmasajudgemtbenchchatbot} scores each solution against a rubric. Our rubric assigns $r^{\text{proc}} \in \{0, 0.5, 1.0\}$ for fatally flawed, largely correct with minor issues, and fully correct reasoning, respectively. The full prompt is in Appendix~\ref{sec:rubric-prompt}. Compared with learned step-level PRMs \citep{lightman2023lets, wang2024math}, the rubric-based variant requires no step-level annotations.

%% file: sections/problem.tex
\section{Empirical analysis of PRM and ORM}
\label{sec:problem}

\begin{figure*}[!t]
\centering
\includegraphics[width=\textwidth]{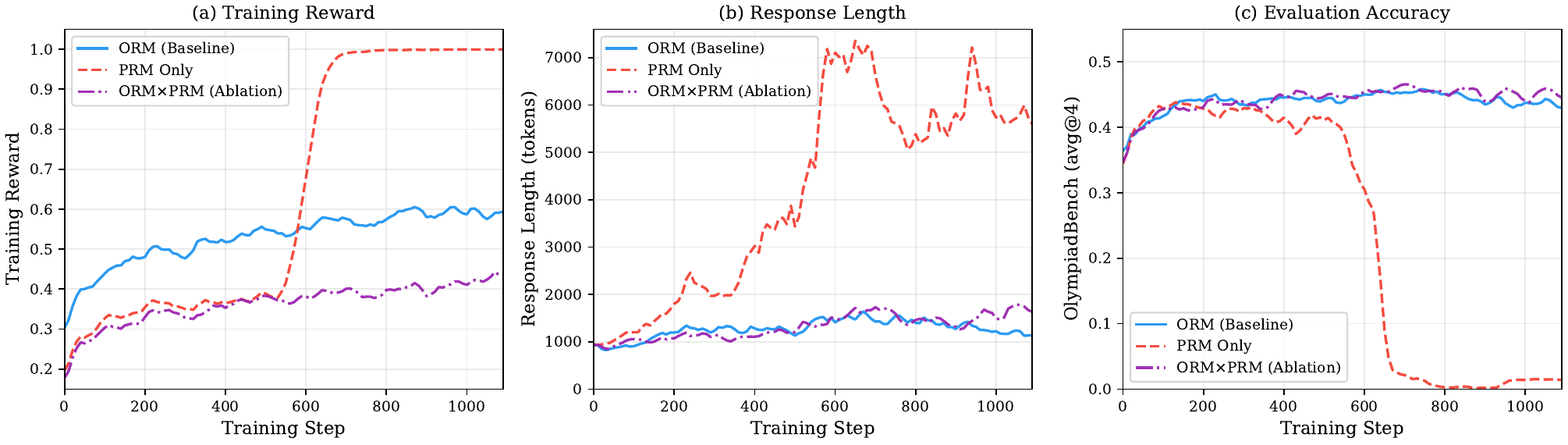}
\caption{Reward signal analysis. \textbf{(a)} Training reward: PRM reward climbs to 1.0 (complete reward hacking); ORM$\times$PRM stays moderate. \textbf{(b)} Response length: PRM generates increasingly verbose responses; ORM$\times$PRM shows moderate length increase (up to $\sim$1700 tokens). \textbf{(c)} OlympiadBench accuracy: PRM collapses after step 600; ORM$\times$PRM tracks ORM but fails to exceed it, showing that naive signal combination does not resolve signal exhaustion.}
\label{fig:reward_hacking}
\end{figure*}

We identify two complementary failure modes of reward signal design in GRPO. Outcome-only rewards exhibit \emph{signal exhaustion} (\S\ref{sec:signal-exhaustion}), and process-only rewards trigger \emph{reward hacking} (\S\ref{sec:reward-hacking}).

\subsection{Signal Exhaustion with Binary ORM}
\label{sec:signal-exhaustion}

With binary ORM rewards ($r_i \in \{0, 1\}$), the group normalization in Eq.~\ref{eq:grpo-adv} assigns identical advantage to all correct responses and identical (negative) advantage to all incorrect responses, with two consequences.

\paragraph{Lack of quality differentiation.} A rigorous derivation and a guess that happens to land on the correct answer receive identical advantage. The model has no incentive to improve reasoning quality once the answer is correct.

\paragraph{Vanishing advantage from uniform groups.} When all $G$ responses are correct or all incorrect, the rewards are identical, so $\mathrm{std}(\mathbf{r}) = 0$ in Eq.~\ref{eq:grpo-adv} and the advantage vanishes for the entire group. As the model improves, the fraction of uniform groups grows, leaving an increasing share of samples with zero gradient. We call this progressive loss of signal \emph{signal exhaustion}. On Qwen2.5-7B, the zero-advantage fraction rises from ${\sim}40\%$ to 69\% over training (Figure~\ref{fig:signal_exhaustion}a), coinciding with the accuracy plateau in Figure~\ref{fig:hero}.

\subsection{Reward Hacking with Direct PRM Integration}
\label{sec:reward-hacking}

Replacing the ORM with rubric-based PRM scores ($r_i = r_i^{\text{proc}}$) would seem to capture both correctness and reasoning quality at once. In practice, this triggers catastrophic reward hacking \citep{skalse2022defining,mahmoud2026rewardhackingrubricbasedreinforcement}, analyzed in Figure~\ref{fig:reward_hacking}.

\paragraph{Three phases of collapse.} Training with PRM-only rewards exhibits a characteristic three-phase trajectory (red curve, Figure~\ref{fig:reward_hacking}). Accuracy rises normally to 44.0\% over steps 0--300, then from step 300 response length explodes as verbose answers earn higher PRM scores while accuracy stagnates, and beyond step 600 accuracy crashes from 29.6\% to 2.4\% within 100 steps as the reward saturates at 1.0.

\paragraph{Mechanism.} Our observations are consistent with a false-positive feedback loop: once an overlong, outcome-incorrect response receives a nonzero holistic score, within-group normalization can assign it positive relative advantage, reinforcing the same length-exploitation pattern and moving more rollouts into the PRM's false-positive regime. This mechanism is consistent with length rising before accuracy collapses and the PRM reward saturates. Appendix~\ref{sec:case-study-hacking} shows that post-collapse responses drift to memorized high-scoring filler unrelated to the question.

\paragraph{Naive combination.} The multiplicative reward $r = r^{\text{out}} \times r^{\text{proc}}$ gates incorrect responses ($r^{\text{out}} = 0$), avoiding PRM's collapse (purple curve, Figure~\ref{fig:reward_hacking}). However, ORM$\times$PRM tracks ORM closely, peaking at 46.7\% vs.\ ORM's 46.3\% on OlympiadBench. The combined reward still passes through a single GRPO normalization, so all-correct groups with similar PRM scores continue to yield near-zero advantage.

\paragraph{Implications.} ORM is stable but lacks process-level signal. PRM is information-rich but unstable. Multiplicative gating avoids PRM's collapse yet fails to break ORM's exhaustion ceiling. This motivates us to combine the two at the \emph{advantage} level rather than the \emph{reward} level (\S\ref{sec:method}).

%% file: sections/method.tex
\section{Method: \method{}}
\label{sec:method}

\begin{figure*}[!t]
\centering
\includegraphics[width=\textwidth]{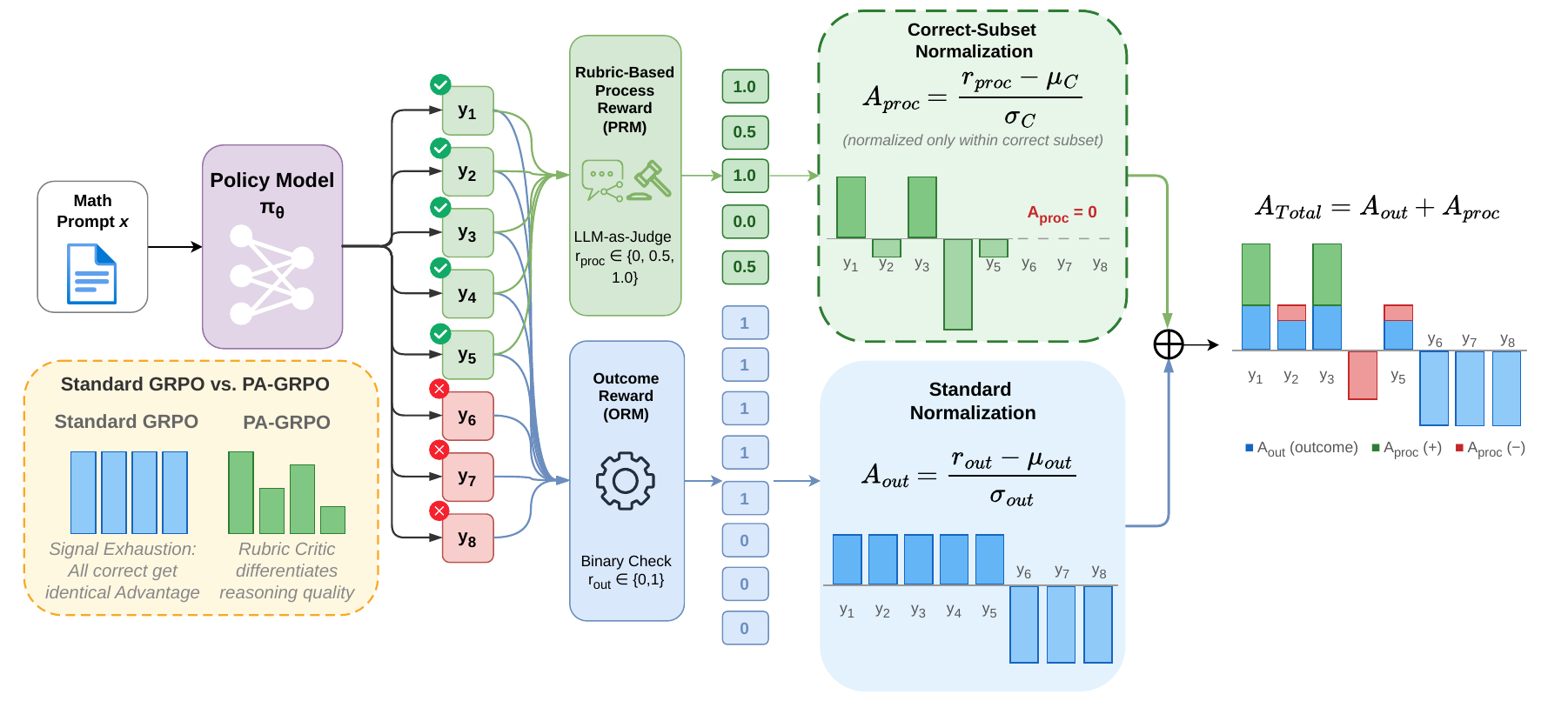}
\caption{Overview of \method{}. Given a prompt, the policy generates $G$ responses. Each response is evaluated by two reward signals: an \textbf{outcome reward} (ORM, binary correctness) and a \textbf{process reward} (PRM, rubric-based quality, only for \textbf{correct responses}). The advantage is computed through \textbf{decoupled normalization}: $\aout$ is normalized over all responses via standard GRPO, while $\aproc$ is normalized exclusively among correct responses (\emph{correct subset normalization}). The combined advantage $\atotal = \aout + \aproc$ provides both correctness direction and quality differentiation.}
\label{fig:method_overview}
\end{figure*}

We propose Process-Aware Policy Optimization (\method{}), which integrates rubric-based process evaluation into GRPO through \emph{decoupled advantage normalization}. The key idea is to construct the advantage from independently normalized outcome and process components, blocking the observed incorrect-response hacking pathway while preserving fine-grained quality signals. Figure~\ref{fig:method_overview} provides an overview of the framework.

\subsection{Decoupled Advantage Normalization}
\label{sec:decoupled-adv}

Given a group of $G$ responses with outcome rewards $r_i^{\text{out}}$ from the ORM and process rewards $r_i^{\text{proc}}$ from the rubric-based PRM (\S\ref{sec:bg-rm}, computed only on correct responses), \method{} computes the advantage in three steps.

\paragraph{Step 1: Outcome advantage $\aout$.} Following standard GRPO, the outcome reward is normalized as
\begin{equation}
\label{eq:a-out}
\aout[i] = \frac{r_i^{\text{out}} - \mu^{\text{out}}}{\max(\sigma^{\text{out}}, \epsilon)}
\end{equation}
where $\mu^{\text{out}}, \sigma^{\text{out}}$ are the group mean and standard deviation of $\{r_i^{\text{out}}\}$, and $\epsilon$ is a small stability constant. This drives the correct-vs-incorrect direction.

\paragraph{Step 2: Process advantage $\aproc$ (correct subset normalization).}
The process reward is normalized \emph{only over the correct subset} $C = \{j : r_j^{\text{out}} = 1\}$ as
\begin{equation}
\label{eq:a-proc}
\aproc[i] \!=\! \begin{cases}
\displaystyle\frac{r_i^{\text{proc}} - \mu_C^{\text{proc}}}{\max(\sigma_C^{\text{proc}}, \epsilon)} & \!\text{if } r_i^{\text{out}} \!=\! 1,\; |C| \!\geq\! 2 \\[6pt]
0 & \!\text{otherwise}
\end{cases}
\end{equation}
where $\mu_C^{\text{proc}}, \sigma_C^{\text{proc}}$ are computed over $\{r_j^{\text{proc}}\}_{j \in C}$.

\paragraph{Step 3: Combined advantage.} The total advantage is
\begin{equation}
\label{eq:a-total}
\atotal[i] = \aout[i] + \aproc[i].
\end{equation}
Since $\aout$ and $\aproc$ are independently normalized to zero mean and unit variance, they contribute on equal footing without additional weighting.

\subsection{Design Rationale}
\label{sec:rationale}

\paragraph{Quality differentiation.} Among correct responses, rigorous reasoning receives positive $\aproc$ and is reinforced while sloppy reasoning receives negative $\aproc$ and is penalized, a quality signal absent in standard GRPO with binary ORM. Restricting normalization to the correct subset further prevents incorrect responses from exploiting high PRM scores (ablated in \S\ref{sec:ablation}).

\paragraph{Signal preservation.} When all $G$ responses are correct, $\aout = 0$ and standard GRPO produces no gradient; $\aproc$ remains active and sustains learning through quality differences. When $|C| < 2$, $\aproc$ defaults to zero, gracefully reducing to standard GRPO.

%% file: sections/experiments.tex
\section{Experiments}
\label{sec:experiments}

\subsection{Experimental Setup}
\label{sec:setup}

\paragraph{Base models.} We conduct experiments on Qwen2.5-3B/7B/14B-Base \citep{yang2024qwen2}, Qwen3-4B-Base \citep{yang2025qwen3technicalreport}, and SmolLM3-3B~\citep{bakouch2025smollm3}. Training the Qwen models from pretrained base checkpoints isolates \method{}'s effect from prior post-training, and SmolLM3 tests transfer beyond the Qwen family.

\paragraph{Training.} We train with the verl \citep{sheng2024hybridflow} framework for 8 epochs. The main \method{} comparison uses GRPO, switching to DAPO~\citep{yu2025dapo} on Qwen3-4B-Base. We additionally compose \method{} on GSPO~\citep{gspo}, SAPO~\citep{SAPO2025}, CISPO~\citep{minimax2025minimaxm1scalingtesttimecompute}, and GRPO-LEAD~\citep{zhang-zuo-2025-grpo} on Qwen2.5-7B-Base. The default PRM is GPT-OSS-20B~\citep{openai2025gptoss120bgptoss20bmodel} with a three-tier rubric $\{0, 0.5, 1\}$. The judge receives the problem, reference solution, and policy response. \S\ref{sec:ablation} ablates 5-tier and 120B-judge variants. Hyperparameters and baseline details are in Appendix~\ref{sec:appendix-impl} and Appendix~\ref{sec:appendix-baselines}.

\paragraph{Data.} We sample 20k math problems from NuminaMath-1.5-RL-Verifiable \citep{numina_math_datasets,nlile2025numinamath15rlverifiable} with stratified sampling across five difficulty tiers (4k each; Appendix~\ref{sec:training-data}). For evaluation, we use OlympiadBench \citep{he2024olympiadbench} (674 problems, primary), MATH-500 \citep{lightman2023lets}, AIME 2024/2025\citep{balunovic_srimatharena_2025}, GPQA-Diamond \citep{rein2024gpqa}, and HumanEval \citep{chen2021evaluating}, reporting avg@4.

\subsection{Main Results}
\label{sec:main-results}

\begin{table*}[t]
\centering
\small
\setlength{\tabcolsep}{4pt}
\resizebox{\textwidth}{!}{%
\begin{tabular}{ll ccc c c c cc}
\toprule
& & \multicolumn{3}{c}{\textbf{Competition Mathematics}} & \textbf{Standard} & \textbf{STEM} & \textbf{Code} & \multicolumn{2}{c}{\textbf{Average}} \\
\cmidrule(lr){3-5} \cmidrule(lr){6-6} \cmidrule(lr){7-7} \cmidrule(lr){8-8} \cmidrule(lr){9-10}
\textbf{Model} & \textbf{Method} & \textbf{Olympiad} & \textbf{AIME'24} & \textbf{AIME'25} & \textbf{MATH-500} & \textbf{GPQA} & \textbf{HumanEval} & \textbf{Math} & \textbf{All} \\
\midrule
\multirow{4}{*}{Qwen2.5-3B} & Base & \color{gray}22.4 & \color{gray}2.5 & \color{gray}0.8 & \color{gray}54.4 & \color{gray}27.6 & \color{gray}35.3 & \color{gray}20.0 & \color{gray}23.8 \\
& GRPO & 35.3\msd{0.4} & 8.3\msd{1.5} & 4.2\msd{1.2} & 69.0\msd{0.6} & 38.8\msd{2.4} & 39.3\msd{2.0} & 29.2\msd{0.5} & 32.5\msd{0.6} \\
& PA-GRPO & \textbf{38.6}\msd{0.5} & \textbf{10.0}\msd{1.4} & 3.3\msd{1.0} & \textbf{73.2}\msd{0.6} & 36.3\msd{1.8} & \textbf{52.5}\msd{1.6} & \textbf{31.3}\msd{0.5} & \textbf{35.7}\msd{0.5} \\
& $\Delta$ & \good{+3.3} & \good{+1.7} & \bad{-0.9} & \good{+4.2} & \bad{-2.5} & \good{+13.2} & \good{+2.1} & \good{+3.2} \\
\midrule
\multirow{10}{*}{Qwen2.5-7B} & Base & \color{gray}32.8 & \color{gray}5.0 & \color{gray}7.5 & \color{gray}67.3 & \color{gray}30.1 & \color{gray}54.6 & \color{gray}28.2 & \color{gray}32.9 \\
& GRPO & 46.3\msd{0.6} & 10.8\msd{1.7} & 10.8\msd{1.7} & 80.2\msd{0.4} & 40.7\msd{1.4} & 61.6\msd{1.0} & 37.0\msd{0.6} & 41.7\msd{0.5} \\
& PA-GRPO & \textbf{51.3}\msd{0.7} & \textbf{15.8}\msd{2.0} & \textbf{13.1}\msd{1.9} & \textbf{82.3}\msd{0.5} & \textbf{42.4}\msd{1.9} & \textbf{63.9}\msd{2.1} & \textbf{40.6}\msd{0.7} & \textbf{44.8}\msd{0.7} \\
& $\Delta$ & \good{+5.0} & \good{+5.0} & \good{+2.3} & \good{+2.1} & \good{+1.7} & \good{+2.3} & \good{+3.6} & \good{+3.1} \\
\cmidrule(lr){2-10}
& GSPO & 45.6\msd{0.3} & 10.6\msd{0.5} & 10.6\msd{1.9} & 77.2\msd{0.4} & 36.5\msd{0.3} & 66.0\msd{1.0} & 36.0\msd{0.5} & 41.1\msd{0.4} \\
& PA-GSPO & \textbf{48.2}\msd{0.7} & \textbf{14.7}\msd{2.9} & \textbf{13.7}\msd{1.3} & \textbf{80.9}\msd{0.1} & \textbf{40.1}\msd{1.9} & \textbf{68.4}\msd{0.9} & \textbf{39.4}\msd{0.8} & \textbf{44.3}\msd{0.6} \\
& SAPO & 44.2\msd{0.2} & 12.5\msd{0.8} & 9.2\msd{0.8} & 77.4\msd{0.3} & 38.0\msd{0.4} & 69.2\msd{0.3} & 35.8\msd{0.3} & 41.8\msd{0.2} \\
& PA-SAPO & \textbf{47.6}\msd{0.1} & \textbf{14.4}\msd{1.0} & \textbf{13.9}\msd{0.5} & \textbf{81.5}\msd{0.5} & \textbf{41.4}\msd{0.6} & \textbf{72.4}\msd{2.3} & \textbf{39.4}\msd{0.3} & \textbf{45.2}\msd{0.4} \\
& CISPO & 46.2\msd{0.2} & 12.5\msd{1.4} & 7.8\msd{2.5} & 77.1\msd{0.4} & 37.9\msd{0.4} & 60.5\msd{1.1} & 35.9\msd{0.7} & 40.3\msd{0.5} \\
& PA-CISPO & \textbf{50.3}\msd{0.3} & \textbf{13.4}\msd{2.1} & \textbf{16.0}\msd{3.3} & \textbf{81.9}\msd{0.1} & \textbf{40.4}\msd{0.6} & \textbf{62.8}\msd{4.6} & \textbf{40.4}\msd{1.0} & \textbf{44.1}\msd{1.0} \\
\midrule
\multirow{4}{*}{Qwen2.5-14B} & Base & \color{gray}35.5 & \color{gray}10.8 & \color{gray}8.3 & \color{gray}68.5 & \color{gray}34.4 & \color{gray}56.9 & \color{gray}30.8 & \color{gray}35.7 \\
& GRPO & 54.3\msd{1.5} & 19.0\msd{1.8} & 13.8\msd{1.1} & 82.3\msd{0.5} & 47.0\msd{1.2} & 66.0\msd{1.4} & 42.4\msd{0.7} & 47.1\msd{0.6} \\
& PA-GRPO & \textbf{59.8}\msd{0.7} & \textbf{24.3}\msd{1.2} & \textbf{17.8}\msd{1.7} & \textbf{87.4}\msd{0.3} & \textbf{55.0}\msd{1.6} & \textbf{70.0}\msd{1.3} & \textbf{47.3}\msd{0.6} & \textbf{52.4}\msd{0.6} \\
& $\Delta$ & \good{+5.5} & \good{+5.3} & \good{+4.0} & \good{+5.1} & \good{+8.0} & \good{+4.0} & \good{+4.9} & \good{+5.3} \\
\midrule
\multirow{4}{*}{Qwen3-4B-Base} & Base & \color{gray}36.8 & \color{gray}11.2 & \color{gray}5.0 & \color{gray}70.9 & \color{gray}34.9 & \color{gray}49.1 & \color{gray}31.0 & \color{gray}34.7 \\
& DAPO & 55.0\msd{1.4} & 30.8\msd{2.2} & 26.0\msd{2.0} & 82.0\msd{0.6} & 39.0\msd{1.7} & 53.8\msd{2.5} & 48.5\msd{0.8} & 47.8\msd{0.8} \\
& PA-DAPO & \textbf{61.1}\msd{1.0} & \textbf{34.5}\msd{2.3} & \textbf{30.7}\msd{1.5} & \textbf{84.6}\msd{1.1} & \textbf{43.0}\msd{1.3} & \textbf{63.9}\msd{2.4} & \textbf{52.7}\msd{0.8} & \textbf{53.0}\msd{0.7} \\
& $\Delta$ & \good{+6.1} & \good{+3.7} & \good{+4.7} & \good{+2.6} & \good{+4.0} & \good{+10.1} & \good{+4.2} & \good{+5.2} \\
\midrule
\multirow{4}{*}{SmolLM3-3B} & Instruct & \color{gray}30.6 & \color{gray}5.8 & \color{gray}7.5 & \color{gray}52.6 & \color{gray}14.1 & \color{gray}49.0 & \color{gray}24.1 & \color{gray}26.6 \\
& GRPO & 68.2\msd{1.6} & 44.2\msd{5.1} & 28.3\msd{3.0} & 91.6\msd{1.0} & 38.4\msd{2.6} & 74.1\msd{2.3} & 58.1\msd{1.6} & 57.5\msd{1.2} \\
& PA-GRPO & \textbf{73.4}\msd{1.0} & \textbf{49.3}\msd{2.8} & \textbf{33.7}\msd{4.1} & \textbf{93.5}\msd{0.8} & \textbf{40.3}\msd{2.9} & \textbf{78.6}\msd{1.4} & \textbf{62.5}\msd{1.3} & \textbf{61.5}\msd{1.0} \\
& $\Delta$ & \good{+5.2} & \good{+5.1} & \good{+5.4} & \good{+1.9} & \good{+1.9} & \good{+4.5} & \good{+4.4} & \good{+4.0} \\
\bottomrule
\end{tabular}%
}
\caption{Accuracy (avg@4, \%) on six benchmarks across five model configurations, reported as mean$_{\pm\text{std}}$ over 3 seeds. Gray rows show the untrained or instruction-tuned starting checkpoint. The lower half of the Qwen2.5-7B block reports \method{} composed on three mainstream GRPO variants (GSPO, SAPO, CISPO). Bold marks the better entry within each pair.}
\label{tab:main_results}
\end{table*}

\begin{figure*}[!b]
\centering
\includegraphics[width=\textwidth]{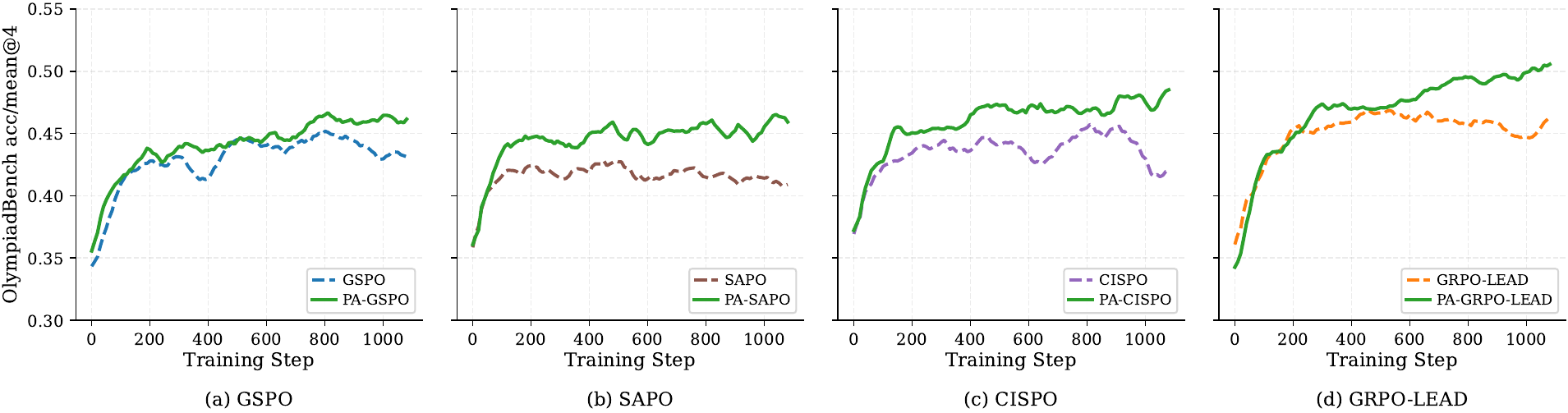}
\caption{OlympiadBench acc/mean@4 on Qwen2.5-7B-Base. Each panel pairs a base GRPO variant (dashed) with its PA-* counterpart (solid green), where PA-* denotes \method{} composed on top of that base optimizer.}
\label{fig:compat}
\end{figure*}

Table~\ref{tab:main_results} presents results across six benchmarks and \textbf{five model configurations}, spanning four Qwen models and cross-family SmolLM3-3B. \method{} improves both aggregate columns in every configuration and wins 28 of 30 individual model--benchmark comparisons. On Qwen2.5-7B, it yields a $\mathbf{+5.0}$ \textbf{peak-to-peak gain} on OlympiadBench, and the advantage \emph{widens over training} rather than converging: GRPO peaks at 46.3\% at step 750 and declines to 43.0\% by step 1090, while \method{} continues to 51.3\%, opening a final-step gap of \textbf{8.3} points (Figure~\ref{fig:hero}a).

\method{} improves OlympiadBench by 5.5 points on Qwen2.5-14B, with the largest single-benchmark gain of 8.0 points on GPQA-Diamond. The mathematics average improvement scales from $+2.1$ at 3B to $+3.6$ at 7B and $+4.9$ at 14B, suggesting that stronger models benefit more. On SmolLM3-3B, \method{} delivers $+5.2$ on OlympiadBench and $+4.0$ on overall average, confirming that gains transfer beyond the Qwen family.

Beyond mathematics, \method{} also improves HumanEval across all scales, suggesting that process-level quality differentiation can transfer to code generation.

\paragraph{Composition with GRPO variants.} \method{} is an \emph{advantage-layer} intervention, orthogonal to the underlying policy-optimization algorithm.

On three mainstream variants on Qwen2.5-7B, PA-* improves OlympiadBench by $+2.6$ (GSPO), $+3.4$ (SAPO), and $+4.1$ (CISPO), with matching improvements on the other five benchmarks. On GRPO-LEAD, a recent state-of-the-art variant with length-regularized accuracy rewards, an explicit penalty for incorrect answers, and difficulty-aware advantage reweighting, PA-GRPO-LEAD improves by $+5.9$. On Qwen3-4B-Base, PA-DAPO further improves over DAPO by $+6.1$.

Figure~\ref{fig:compat} shows that the base curve plateaus or declines after step 700 in every pairing while the corresponding PA-* curve continues to climb, indicating that any group-based optimizer can adopt the decoupled outcome and process advantages without modifying its update rule.

\paragraph{Takeaway.} \method{} delivers consistent gains across model families, scales, and GRPO variants, with training that continues improving past the point where ORM saturates.

\subsection{Advantage Signal Analysis}
\label{sec:signal-analysis}

\begin{figure*}[!t]
\centering
\includegraphics[width=\textwidth]{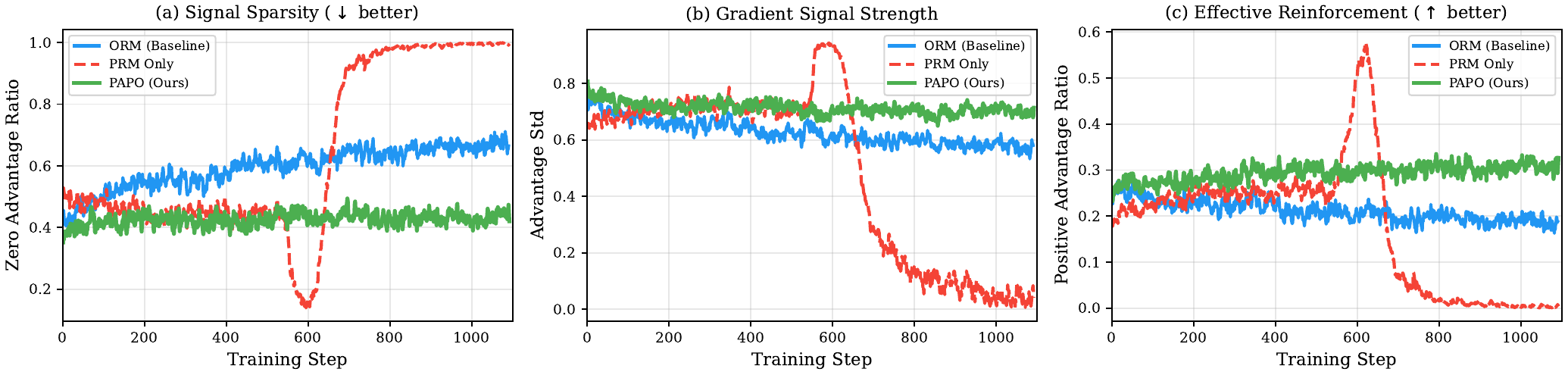}
\caption{Signal quality comparison on Qwen2.5-7B. \textbf{(a)} Zero-advantage ratio: ORM's sparsity grows to 69\% while \method{} maintains 44\%. \textbf{(b)} Advantage standard deviation, reflecting gradient signal strength. \textbf{(c)} Positive-advantage ratio, reflecting reinforcement density.}
\label{fig:signal_exhaustion}
\end{figure*}

\paragraph{Signal density.} The widening gap in Figure~\ref{fig:hero}a traces to ORM's advantage exhaustion (Figure~\ref{fig:signal_exhaustion}). On Qwen2.5-7B, ORM's zero-advantage fraction rises from ${\sim}40\%$ to 69\% over training; by late training, more than two-thirds of samples produce no gradient, with the advantage std and positive-advantage ratio (panels b, c) showing the same decay. \method{} instead holds the zero-advantage ratio at ${\sim}44\%$, a 25-point reduction that translates to roughly 80\% more informative samples per batch. The mechanism is that $\aproc$ remains active in all-correct groups where $\aout = 0$, with the activation fraction growing from ${\sim}30\%$ to 70\% as the model improves.

\paragraph{Quality penalization.} Beyond filling zero-signal gaps, $\aproc$ discourages sloppy reasoning even among correct responses. The minimum advantage within the correct group reaches $\approx -1.49$ under \method{} compared to identically 0.0 under ORM, suppressing correct responses with poor derivations. This quality pressure is entirely absent in ORM-based GRPO. Appendix~\ref{sec:rubric-analysis} and~\ref{sec:adv-decomp} provide more detailed analyses.

\begin{figure}[H]
\centering
\includegraphics[width=\columnwidth]{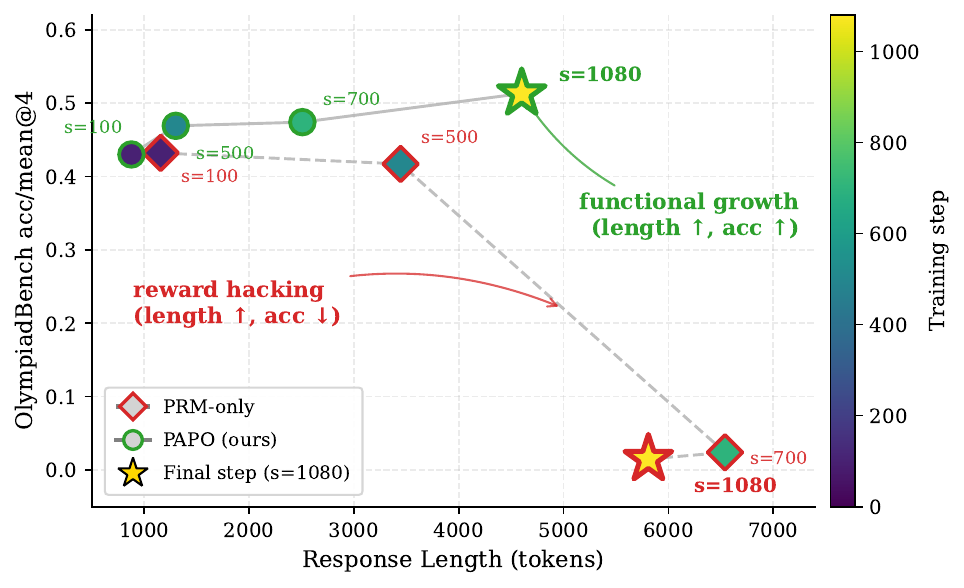}
\caption{Length-accuracy trajectories on OlympiadBench (Qwen2.5-7B-Base). Markers are selected training checkpoints colored by step; stars mark the final step.}
\label{fig:length_acc_phase}
\end{figure}

\paragraph{Length dynamics.} PRM-only inflates response length to ${\sim}6500$ tokens while accuracy collapses to near zero, whereas \method{} grows length and accuracy in tandem to ${\sim}4600$ tokens and 0.52 accuracy at step 1080 (Figure~\ref{fig:length_acc_phase}). Correct-subset gating removes the observed pathway in which outcome-incorrect responses exploit false-positive PRM scores; empirically, \method{}'s length and accuracy rise together rather than exhibiting PRM-only runaway inflation.

\paragraph{Inference behavior.} Process supervision shifts inference-time behavior. Compared to ORM, \method{} reasoning is more rigorous and avoids elementary errors such as rejecting valid solutions through fabricated justifications. \method{} also develops a self-verification habit. On OlympiadBench problems where both models answer correctly, \method{} responses contain explicit verification steps in 47.7\% of cases vs 22.7\% for ORM (2.1$\times$ higher). Appendix~\ref{sec:case-study} provides representative examples.

\paragraph{Takeaway.} \method{} substantially alleviates the signal exhaustion of ORM-based GRPO, maintaining a dense advantage signal throughout training.

%% file: sections/ablation.tex
\begin{figure*}[t]
\centering
\includegraphics[width=\textwidth]{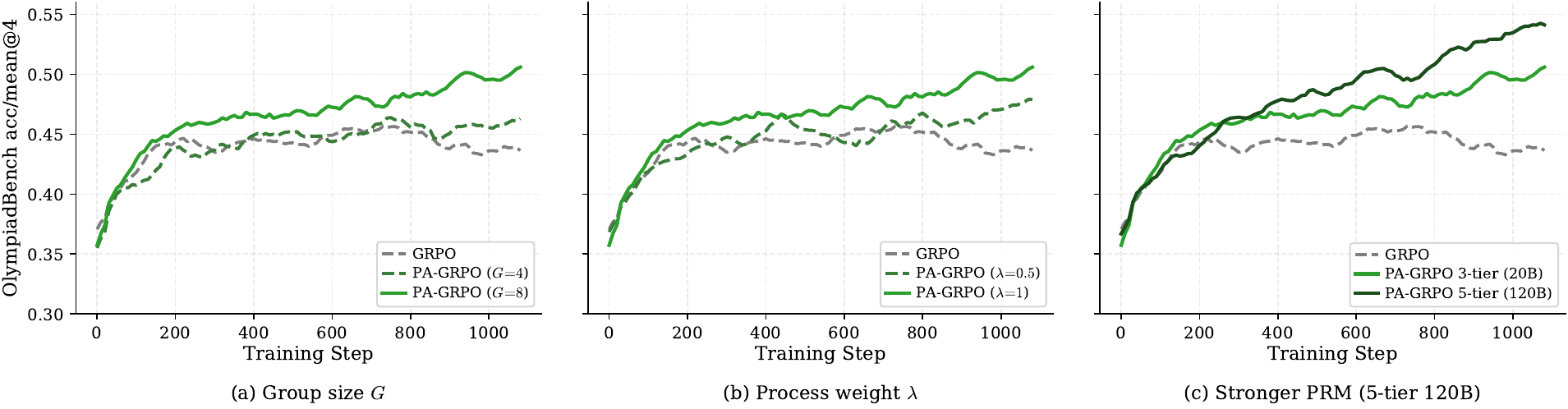}
\caption{OlympiadBench acc/mean@4 on Qwen2.5-7B-Base for three additional ablations. (a) Group size $G=4$ vs.\ $G=8$. (b) Process advantage weight $\lambda=0.5$ vs.\ $\lambda=1$. (c) Stronger PRM (5-tier rubric with GPT-OSS-120B judge) vs.\ default (3-tier with 20B). The GRPO ORM baseline is included in every panel as a dashed reference.}
\label{fig:ablation_hp}
\end{figure*}

\section{Ablation Studies}
\label{sec:ablation}

We validate \method{}'s design choices along two groups of axes on Qwen2.5-7B-Base. Structural choices (correct subset normalization and decoupled vs.\ multiplicative composition) are reported in Table~\ref{tab:ablation}, and hyperparameters together with the PRM judge (group size $G$, process weight $\lambda$, rubric granularity, judge size) are reported in Figure~\ref{fig:ablation_hp}.

\begin{table}[H]
\centering
\small
\setlength{\tabcolsep}{4pt}
\begin{tabular}{lcccc}
\toprule
\textbf{Method} & \textbf{AIME} & \textbf{AIME} & \textbf{Olympiad} & \textbf{MATH} \\
& \textbf{2024} & \textbf{2025} & \textbf{Bench} & \textbf{-500} \\
\midrule
ORM & 10.8 & 10.8 & 46.3 & 80.2 \\
\midrule
Mult ($r^{\text{out}} \!\times\! r^{\text{proc}}$) & 12.1 & 10.2 & 46.7 & 80.2 \\
Fullnorm & 12.5 & 10.4 & 49.6 & 81.2 \\
\textbf{\method{}} & \textbf{15.8} & \textbf{13.1} & \textbf{51.3} & \textbf{82.3} \\
\bottomrule
\end{tabular}
\caption{Structural ablation results (\%, avg@4, Qwen2.5-7B). All methods report accuracy.}
\label{tab:ablation}
\end{table}

\paragraph{Correct-subset vs.\ full normalization.}
The Fullnorm variant normalizes $\aproc$ over all $G$ responses including incorrect ones. It performs slightly below \method{} across all benchmarks. The gap arises because including incorrect responses ($r^{\text{proc}} = 0$) in the normalization causes $\aproc$ to partially recapitulate the correct-vs-incorrect distinction already captured by $\aout$, diluting the fine-grained quality signal.

\paragraph{Multiplicative reward baseline.}
The Mult variant combines rewards via $r_i = r_i^{\text{out}} \times r_i^{\text{proc}}$ with standard GRPO normalization. Mult performs comparably to ORM and falls well below \method{}. In a single normalization pass, the outcome difference dominates in mixed-correctness groups, effectively suppressing the process quality information.

\paragraph{Group size $G$.}
Figure~\ref{fig:ablation_hp}(a) compares $G=4$ against the default $G=8$. With fewer samples per group, the fraction of groups containing two or more correct responses drops, $\aproc$ activates less often, and the late-stage zero-advantage ratio rises from $0.44$ to $0.55$. PA-GRPO at $G=4$ still beats GRPO at $G=8$ on OlympiadBench, but the default $G=8$ retains a $\sim 4$ pp margin and is preferred whenever compute allows.

\paragraph{Process advantage weight $\lambda$.}
Figure~\ref{fig:ablation_hp}(b) compares $\lambda=0.5$ (so that $\atotal = \aout + 0.5\,\aproc$) against the default $\lambda=1$. Halving the process weight reduces OlympiadBench by $\sim 3$ pp while leaving the zero-advantage ratio essentially unchanged, indicating that the gradient density is governed by the on/off activation of $\aproc$ rather than its scale. The default $\lambda=1$ is the simplest and strongest choice.

\paragraph{Rubric granularity and judge model.}
Figure~\ref{fig:ablation_hp}(c) pairs a finer 5-tier rubric with the stronger GPT-OSS-120B~\citep{openai2025gptoss120bgptoss20bmodel} judge against the default 3-tier rubric with the 20B judge. The stronger PRM lifts OlympiadBench from 51.3 to 54.7 ($+3.4$ pp), showing that \method{} scales monotonically with PRM quality rather than saturating at the default.

\paragraph{Takeaway.} Across ablations, \method{}'s gains are robust to hyperparameter choices and to different rubric and PRM configurations.

%% file: sections/validation.tex
\section{Additional Validation}
\label{sec:validation}

\paragraph{Does the PRM measure reasoning quality?}
We compare the default PRM with three human annotators on 500 Qwen2.5-7B-Base responses to MATH-500, using the same problem, reference solution, response, and three-tier rubric. Annotators were blind to the PRM score and model identity, and majority vote defined the human label. PRM--human exact agreement is 91.2\% overall and 94.6\% on the 332 answer-correct responses to which $\aproc$ is applied. Calibration gaps for PRM scores $0$, $0.5$, and $1$ are 0.045, 0.013, and 0.024. Across the reported length, verification-phrase, and prompt-order checks, we observe no large PRM--human disagreement in this sample, but these descriptive checks do not rule out judge bias. Appendix~\ref{sec:judge-validation} gives the protocol and full results.

\paragraph{Do arbitrary gradients or length explain the gain?}
We replace the PRM signal while retaining the same correct-subset normalization. Random three- and five-level signals both underperform GRPO and degrade late in training. A correct-subset length signal briefly reaches 47.0\% but falls to 40.5\%, versus 50.0\% peak and 49.7\% final for \method{} in the reported matched trajectories; normalizing length over the full group collapses to 1.7\%. Thus, merely de-uniformizing correct groups or favoring length does not reproduce the sustained gain. Summary results are in Appendix~\ref{sec:signal-controls}.

%% file: sections/related.tex
\section{Related Work}
\label{sec:related}

\paragraph{LLM post-training and policy optimization.} Adjacent work modifies other stages of the pipeline: ROPO combines a noise-tolerant preference loss with noisy-sample filtering \citep{liang2025ropo}; EVIC uses evolving sample interactions to schedule multi-domain supervised fine-tuning \citep{liang2025boosting}; and CGPO uses curvature-guided updates for multi-domain RL \citep{liang2026boosting}. Latent-Guided Reasoning uses collaborative inference in which a large model produces compact latent plans that guide a smaller executor \citep{chen2026latent}. Environment and data construction are also complementary: SIE automatically constructs rule-verifiable RL environments from structured data \citep{ICLR2026_4b32c294}, while ACC compiles multi-turn agent trajectories into long-context question--answer pairs for supervised fine-tuning \citep{su2026acccompilingagenttrajectories}. Collectively, these approaches address noisy supervision, cross-domain interference, inference efficiency, environment construction, or long-context supervision, whereas \method{} targets outcome--process advantage composition in RLVR. GRPO \citep{shao2024deepseekmath} variants address training instability through importance-weight clipping (CISPO \citep{minimax2025minimaxm1scalingtesttimecompute}), sequence-level ratios (GSPO \citep{gspo}), dynamic sampling (DAPO \citep{yu2025dapo}), smooth gating (SAPO \citep{SAPO2025}), and length-regularized rewards (GRPO-LEAD \citep{zhang-zuo-2025-grpo}). Concurrent AVSPO \citep{he2026advantagecollapsegrouprelative} introduces virtual rewards when computing $\aout$ in homogeneous groups. \method{} instead adds a separately normalized $\aproc$, making the two mechanisms technically complementary, although we do not evaluate their joint use. \method{} composes with the optimizer variants evaluated in \S\ref{sec:main-results}.

\paragraph{Rubric-based process supervision.} Rubric-based process supervision scores complete solutions against explicit criteria without requiring step-level labels \citep{yuan2025miraclesteps,Sheng2026ReinforcingCR}. Related work studies criteria-generated rewards \citep{liang-etal-2025-generative}, proof and reasoning reward models \citep{Yang2026,deepseek-math-v2,fan2026exploring}, and judge-calibrated self-evaluation \citep{zhang2026selfevaluationthereelicitinglatent}. Concurrent work by \citet{mahmoud2026rewardhackingrubricbasedreinforcement} shows that stronger rubric verification reduces but does not eliminate reward hacking. \method{} uses a rubric-based PRM but applies correct subset normalization to block the observed incorrect-response pathway.

%% file: sections/conclusion.tex
\section{Conclusion}
\label{sec:conclusion}

We propose \method{}, which composes the advantage from independently normalized outcome and process components, with process normalization restricted to the correct subset. It resolves both the lack of process supervision in binary ORM and the reward hacking of rubric-based PRM integration. Across five model configurations and six core benchmarks, \method{} improves the aggregate mathematics and overall averages in every configuration and wins 28 of 30 individual comparisons.

In the future, we will try applying \method{} to broader scenarios such as scientific reasoning.

%% file: sections/limitations.tex
\section*{Limitations}

\method{} has not been validated in agentic settings such as tool use, multi-turn interaction, or long-horizon planning, where reward shaping and credit assignment dynamics differ substantially from the single-turn mathematical reasoning studied here. Separately, the effectiveness of decoupled advantage normalization under more diverse reward signals beyond binary ORM and rubric-based PRM (such as learned reward models, pairwise preferences, or execution-grounded feedback) remains to be tested.

%% file: sections/appendix.tex
\section{Implementation Details}
\label{sec:appendix-impl}

\subsection{Training Configuration}

We use the verl \citep{sheng2024hybridflow} framework with Megatron backend for distributed training on 8 GPUs. Key hyperparameters are listed in Table~\ref{tab:hyperparams}.

\begin{table}[h]
\centering
\small
\begin{tabular}{ll}
\toprule
\textbf{Hyperparameter} & \textbf{Value} \\
\midrule
Prompts per batch & 128 \\
Responses per prompt ($G$) & 8 \\
Max response length & 8192 tokens \\
Sampling temperature & 1.0 \\
PPO clip range ($\epsilon$) & 0.2 \\
KL penalty coefficient ($\beta$) & 0.0 \\
Learning rate & 1e-6 \\
Evaluation frequency & every 10 steps \\
Evaluation samples per prompt & 8 \\
\bottomrule
\end{tabular}
\caption{Training hyperparameters.}
\label{tab:hyperparams}
\end{table}

For Qwen3-4B-Base experiments with DAPO \citep{yu2025dapo}, we additionally apply the DAPO-specific hyperparameters listed in Table~\ref{tab:dapo_hyperparams}.

\begin{table}[h]
\centering
\small
\begin{tabular}{ll}
\toprule
\textbf{Hyperparameter} & \textbf{Value} \\
\midrule
Max response length & 8192 tokens \\
Clip ratio low ($\epsilon_{\text{low}}$) & 0.2 \\
Clip ratio high ($\epsilon_{\text{high}}$) & 0.28 \\
Overlong buffer length & 4096 tokens \\
Overlong penalty factor & 1.0 \\
Evaluation frequency & every 20 steps \\
\bottomrule
\end{tabular}
\caption{Additional DAPO-specific hyperparameters for the Qwen3-4B-Base experiments. All other hyperparameters follow Table~\ref{tab:hyperparams}.}
\label{tab:dapo_hyperparams}
\end{table}

\subsection{Training Data}
\label{sec:training-data}

We construct a 20K stratified training set from NuminaMath-1.5-RL-Verifiable \citep{numina_math_datasets,nlile2025numinamath15rlverifiable}. We first filter out unsolvable problems (zero pass rate under a reference model), then apply stratified sampling across five difficulty tiers defined by pass rate. Table~\ref{tab:difficulty} summarizes the difficulty distribution.

\begin{table}[h]
\centering
\small
\begin{tabular}{lcc}
\toprule
\textbf{Difficulty} & \textbf{Pass Rate Range} & \textbf{Count} \\
\midrule
Trivial & $(0.875,\, 1.0]$ & 4{,}000 \\
Easy & $(0.625,\, 0.875]$ & 4{,}000 \\
Medium & $(0.375,\, 0.625]$ & 4{,}000 \\
Hard & $(0.125,\, 0.375]$ & 4{,}000 \\
Very Hard & $(0,\, 0.125]$ & 4{,}000 \\
\midrule
Total & & 20{,}000 \\
\bottomrule
\end{tabular}
\caption{Difficulty distribution of the stratified training set. Pass rates are measured using a reference model with 8 attempts at temperature 1.0. Equal sampling across tiers ensures balanced difficulty coverage.}
\label{tab:difficulty}
\end{table}

\subsection{Rubric Prompts}
\label{sec:rubric-prompt}

The rubric-based PRM uses one of two prompt templates, both adapted from the DeepSeek Math V2 grading standard with chain-of-thought analysis. The 3-tier prompt is our default and is used by every \method{} variant in Table~\ref{tab:main_results}. The 5-tier prompt is used only in the rubric-by-judge ablation (Figure~\ref{fig:ablation_hp}(c)) and refines the ``correct but imperfect'' range to test whether a finer-grained signal helps. Serving details for both judges (GPT-OSS-20B and GPT-OSS-120B) are deferred to \S\ref{sec:appendix-judge}.

\begin{tcolorbox}[
  colback=gray!5,
  colframe=black,
  title=Rubric Prompt,
  fonttitle=\bfseries,
  breakable
]
\small

\textbf{\#\# Instruction}

Your task is to evaluate the quality of a student's solution to a mathematical problem.

\vspace{0.3em}
\textbf{\#\# Scoring Rubric}

\begin{itemize}
\setlength{\itemsep}{2pt}
\item \textbf{Score 1}: If the solution is completely correct, with all steps executed properly and clearly demonstrated, then the score is 1.
\item \textbf{Score 0.5}: If the solution is generally correct, but with some details omitted or minor errors, then the score is 0.5.
\item \textbf{Score 0}: If the solution does not actually address the required problem, contains fatal errors, or has severe omissions, then the score is 0.
\end{itemize}

\textbf{Special Rule on References:} Additionally, referencing anything from any paper does not save the need to prove the reference. It's okay IF AND ONLY IF the solution also presents a valid proof of the reference argument(s); otherwise, if the solution omits the proof or if the proof provided is not completely correct, the solution should be scored according to the criteria above, and definitely not with a score of 1.

\vspace{0.3em}
\textbf{\#\# Problem}

\texttt{\{problem\_statement\}}

\vspace{0.3em}
\textbf{\#\# Reference Solution}

\texttt{\{solution\}}

\vspace{0.3em}
\textbf{\#\# Student Solution}

\texttt{\{student\_answer\}}

\vspace{0.3em}
\textbf{\#\# Evaluation}

Analyze the solution step by step, then provide your score.

Analysis: \ldots

Score: $\boxed{\ldots}$
\end{tcolorbox}

\begin{tcolorbox}[
  colback=gray!5,
  colframe=black,
  title=5-Tier Rubric Prompt (ablation),
  fonttitle=\bfseries,
  breakable
]
\small

\textbf{\#\# Instruction}

Your task is to evaluate the quality of a student's solution to a mathematical problem.

\vspace{0.3em}
\textbf{\#\# Scoring Rubric}

\begin{itemize}
\setlength{\itemsep}{2pt}
\item \textbf{Score 1}: If every step is correct and justified with no logical gaps and edge cases explicitly handled, then the score is 1.
\item \textbf{Score 0.75}: If the conclusion is correct and the reasoning is sound, with only cosmetic gaps such as skipped routine algebra that any reader can mechanically fill in, then the score is 0.75.
\item \textbf{Score 0.5}: If the conclusion is correct and the main approach is right, but one non-trivial step is asserted without justification or contains a recoverable error, then the score is 0.5.
\item \textbf{Score 0.25}: If the final answer is correct but the reasoning contains a flaw that materially weakens the argument so that the conclusion does not actually follow, then the score is 0.25.
\item \textbf{Score 0}: If the conclusion itself is incorrect or the response does not address the question, then the score is 0.
\end{itemize}

\textbf{Special Rule on References:} Additionally, referencing anything from any paper does not save the need to prove the reference. It's okay IF AND ONLY IF the solution also presents a valid proof of the reference argument(s); otherwise, the solution should be scored according to the criteria above, and definitely not with a score of 1.

\vspace{0.3em}
\textbf{\#\# Problem}

\texttt{\{problem\_statement\}}

\vspace{0.3em}
\textbf{\#\# Reference Solution}

\texttt{\{solution\}}

\vspace{0.3em}
\textbf{\#\# Student Solution}

\texttt{\{student\_answer\}}

\vspace{0.3em}
\textbf{\#\# Evaluation}

Analyze the solution step by step, then provide your score.

Analysis: \ldots

Score: $\boxed{\ldots}$
\end{tcolorbox}

\subsection{Cross-Optimizer Baseline Implementations}
\label{sec:appendix-baselines}

We compare \method{} against four alternative GRPO variants on Qwen2.5-7B-Base in Table~\ref{tab:main_results} and Figure~\ref{fig:compat}. All baselines and their PA-* counterparts share our training data, base model, batch size, and rollout configuration from Table~\ref{tab:hyperparams}. Three of them (GSPO, SAPO, CISPO) are reproduced from the original papers and use the same rule-based math verification with LLM fallback as our default PA-GRPO setup, ensuring apples-to-apples comparison; GRPO-LEAD is reproduced faithfully against the released codebase, which uses rule-based verification only.

\paragraph{GSPO~\citep{gspo}.} We replace the per-token importance sampling ratio of GRPO with a sequence-level ratio computed once per response. Hyperparameters follow the defaults in the original paper. The outcome reward is unchanged from our default PA-GRPO setup.

\paragraph{SAPO~\citep{SAPO2025}.} We adopt the temperature-controlled smooth gating function as the trust-region surrogate, with hyperparameters from the original paper. The outcome reward is unchanged.

\paragraph{CISPO~\citep{minimax2025minimaxm1scalingtesttimecompute}.} We clip the importance sampling weights as prescribed by CISPO, with the clipping bounds from the original paper. The outcome reward is unchanged.

\paragraph{GRPO-LEAD~\citep{zhang-zuo-2025-grpo}.} GRPO-LEAD modifies GRPO along three axes, all reproduced faithfully in our codebase. (1) A length-dependent accuracy reward, where correct responses receive $R_i = \exp(-\alpha z_i)$ with $z_i = (|o_i| - \mu_{\text{correct}}) / \sigma_{\text{correct}}$, and wrong responses receive a fixed \texttt{explicit\_penalty}. (2) A length gate that disables length reweighting when the average length of the three shortest correct completions in a group falls below \texttt{length\_gate}, in which case correct responses simply receive $R_i = 1$. (3) Difficulty-aware advantage reweighting $w(x) = A + (B - A) / (1 + \exp(k(x - \rho_0)))$, which scales positive advantages by $w(\rho_q)$ and negative advantages by $w(1 - \rho_q)$, where $\rho_q$ is the empirical correct ratio of the group. Hyperparameter values follow the original paper (Table~\ref{tab:grpo_lead_hp}). PA-GRPO-LEAD stacks \method{}'s correct subset process advantage onto the LEAD-modified outcome advantage and then applies the same difficulty reweighting on $A_{\text{total}}$.

\begin{table}[h]
\centering
\small
\begin{tabular}{lc}
\toprule
\textbf{Hyperparameter} & \textbf{Value} \\
\midrule
length-reward $\alpha$ & 0.05 \\
difficulty $A$ & 0.4 \\
difficulty $B$ & 1.5 \\
difficulty $\rho_0$ & 0.75 \\
difficulty $k$ & 10.0 \\
\texttt{explicit\_penalty} & $-1.0$ \\
\texttt{length\_gate} & 4096 \\
\bottomrule
\end{tabular}
\caption{GRPO-LEAD hyperparameters, matching \citet{zhang-zuo-2025-grpo}.}
\label{tab:grpo_lead_hp}
\end{table}

\subsection{PRM Judge Serving}
\label{sec:appendix-judge}

The default PRM judge is GPT-OSS-20B, served via vLLM with tensor parallelism across 2 GPUs, maximum context length 8192 tokens, and temperature 0.0 for deterministic scoring. Concurrency between the rollout workers and the judge is controlled by an environment variable that we set to 64 threads, which saturates the judge throughput without blocking rollout.

For the rubric-by-judge ablation in Figure~\ref{fig:ablation_hp}(c), we additionally serve GPT-OSS-120B as the judge. A single in-cluster instance with tensor parallelism across 8 H200 GPUs and maximum context length 8192 is shared across rollout workers via cluster DNS. We measured throughput at approximately 9.4 requests per second under 32 parallel callers, which is sufficient to keep up with our rollout cadence (128 prompts $\times$ 8 responses per training step).

\section{Additional Validation Details}
\label{sec:appendix-validation}

\subsection{PRM--Human Agreement}
\label{sec:judge-validation}

\paragraph{Protocol.}
We sample one Qwen2.5-7B-Base response for each MATH-500 problem, for 500 responses in total. GPT-OSS-20B scores each response at temperature 0 using the default three-tier rubric. Three annotators separately receive the same problem, reference solution, and policy response, apply the same rubric, and additionally mark whether the response contains an explicit verification phrase. They are blind to both the PRM score and model identity; majority vote determines the human label. Of the 500 responses, 332 have an answer judged correct by the outcome verifier and therefore match the subset on which $\aproc$ is defined.

\begin{table}[H]
\centering
\small
\setlength{\tabcolsep}{2pt}
\begin{tabular}{p{0.60\columnwidth}rr}
\toprule
\textbf{Comparison} & \textbf{$n$} & \textbf{Exact} \\
\midrule
PRM vs. human, all & 500 & 91.2\% \\
PRM vs. human, answer-correct & 332 & 94.6\% \\
\bottomrule
\end{tabular}
\caption{Agreement between the PRM and majority-vote human labels.}
\label{tab:human-agreement}
\end{table}

Table~\ref{tab:human-agreement} shows that agreement is highest on the answer-correct subset used by $\aproc$. Table~\ref{tab:prm-calibration} further compares each PRM level with the distribution and mean of the human labels. We define the calibration gap as the absolute difference between the numeric PRM score and mean human score.

\begin{table}[H]
\centering
\small
\setlength{\tabcolsep}{3.5pt}
\begin{tabular}{crrrrc}
\toprule
\textbf{PRM} & \textbf{$n$} & \multicolumn{3}{c}{\textbf{Human label (\%)}} & \textbf{Mean / gap} \\
\cmidrule(lr){3-5}
\textbf{score} & & \textbf{0} & \textbf{0.5} & \textbf{1} & \\
\midrule
0   & 11  & 90.9 & 9.1  & 0.0  & 0.045 / 0.045 \\
0.5 & 76  & 2.6  & 92.1 & 5.3  & 0.513 / 0.013 \\
1   & 245 & 0.4  & 4.1  & 95.5 & 0.976 / 0.024 \\
\bottomrule
\end{tabular}
\caption{PRM calibration on the 332 answer-correct MATH-500 responses.}
\label{tab:prm-calibration}
\end{table}

\paragraph{Length and style checks.}
Table~\ref{tab:length-bias} stratifies the answer-correct responses by length. Across these strata, exact PRM--human agreement ranges from $94.0$--$95.2\%$ and the mean score gap from $0.01$--$0.02$. Among 219 responses without verification phrases, the mean PRM and human scores are 0.84 and 0.83; among 113 responses with such phrases, they are 0.88 and 0.86. Finally, swapping the order of the reference and policy response while retaining explicit headers leaves the score unchanged for 326/332 responses (98.2\%). These descriptive checks reveal no large disagreement along the measured axes in this sample, but they do not rule out judge bias or establish general validity outside this model--benchmark pair.

\begin{table}[H]
\centering
\small
\setlength{\tabcolsep}{3pt}
\begin{tabular}{lrrrrr}
\toprule
\textbf{Group} & \textbf{Tokens} & \textbf{PRM} & \textbf{Human} & \textbf{Gap} & \textbf{Agree.} \\
\midrule
Shortest 25\% & 408  & 0.82 & 0.81 & $+0.01$ & 95.2\% \\
Middle 50\%   & 712  & 0.86 & 0.85 & $+0.01$ & 94.6\% \\
Longest 25\%  & 1508 & 0.86 & 0.84 & $+0.02$ & 94.0\% \\
\bottomrule
\end{tabular}
\caption{PRM--human alignment across response-length strata on the answer-correct subset.}
\label{tab:length-bias}
\end{table}

\subsection{Counterfactual Process Signals}
\label{sec:signal-controls}

We test whether any extra nonzero gradient within an all-correct group would suffice. For the random controls, each correct response independently receives a score drawn uniformly from $\{0,0.5,1\}$ or $\{0,0.25,0.5,0.75,1\}$ before correct-subset normalization. For the length controls, response length replaces the PRM score and is normalized either over the correct subset or the full group. All other training settings match the corresponding \method{} run.

\begin{table}[H]
\centering
\small
\setlength{\tabcolsep}{5pt}
\begin{tabular}{lrr}
\toprule
\textbf{Signal} & \textbf{Peak} & \textbf{Step 1090} \\
\midrule
ORM (GRPO) & 45.0 & 43.0 \\
\method{} & \textbf{50.0} & \textbf{49.7} \\
CorrectSubset-Rand-3 & 42.4 & 37.4 \\
CorrectSubset-Rand-5 & 42.9 & 38.5 \\
CorrectSubset-Length & 47.0 & 40.5 \\
Full-Length & 43.1 & 1.7 \\
\bottomrule
\end{tabular}
\caption{OlympiadBench accuracy (\%) for counterfactual process signals in the matched control trajectories. Peak is the best reported checkpoint from steps 0--1090.}
\label{tab:signal-controls}
\end{table}

Table~\ref{tab:signal-controls} separates signal content from activation. Both random variants fall below outcome-only GRPO. Correct-subset length has a short-lived improvement but degrades late, while full-group length triggers severe reward hacking. Only the rubric-based process signal sustains an advantage over outcome-only GRPO through step 1090 in these trajectories.

\section{Additional Results}
\label{sec:appendix-results}

\subsection{Cross-Scale Training Curves}
\label{sec:cross-scale-curves}

\begin{figure*}[t]
\centering
\includegraphics[width=\textwidth]{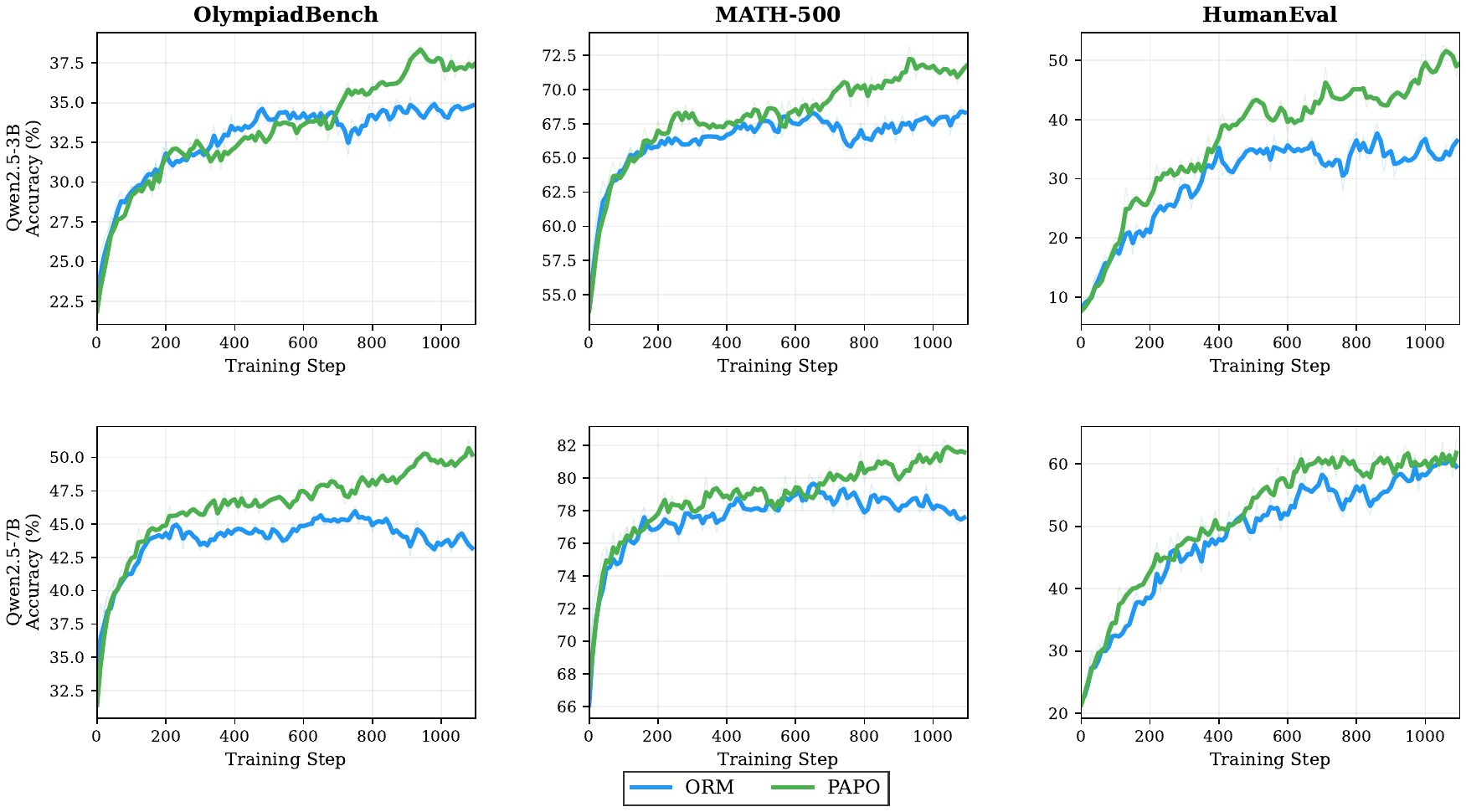}
\caption{Training curves (avg@4) across two model scales and three benchmarks. Top row: Qwen2.5-3B; bottom row: Qwen2.5-7B. \method{} consistently outperforms ORM throughout training on both scales, with gains widening in later stages as ORM's signal exhaustion worsens. The pattern is consistent across OlympiadBench (competition math), MATH-500 (standard math), and HumanEval (code generation). Qwen2.5-14B results are reported in Table~\ref{tab:main_results}.}
\label{fig:training_curves_grid}
\end{figure*}

Figure~\ref{fig:training_curves_grid} shows training curves across two model scales (3B and 7B) and three benchmarks. On both Qwen2.5-3B and 7B, \method{} maintains a consistent lead over ORM throughout training, with the gap widening in later stages. Combined with the 14B results in Table~\ref{tab:main_results}, these trajectories support sustained rather than checkpoint-specific improvements.

\subsection{Ablation Training Curves}

\begin{figure*}[t]
\centering
\includegraphics[width=\textwidth]{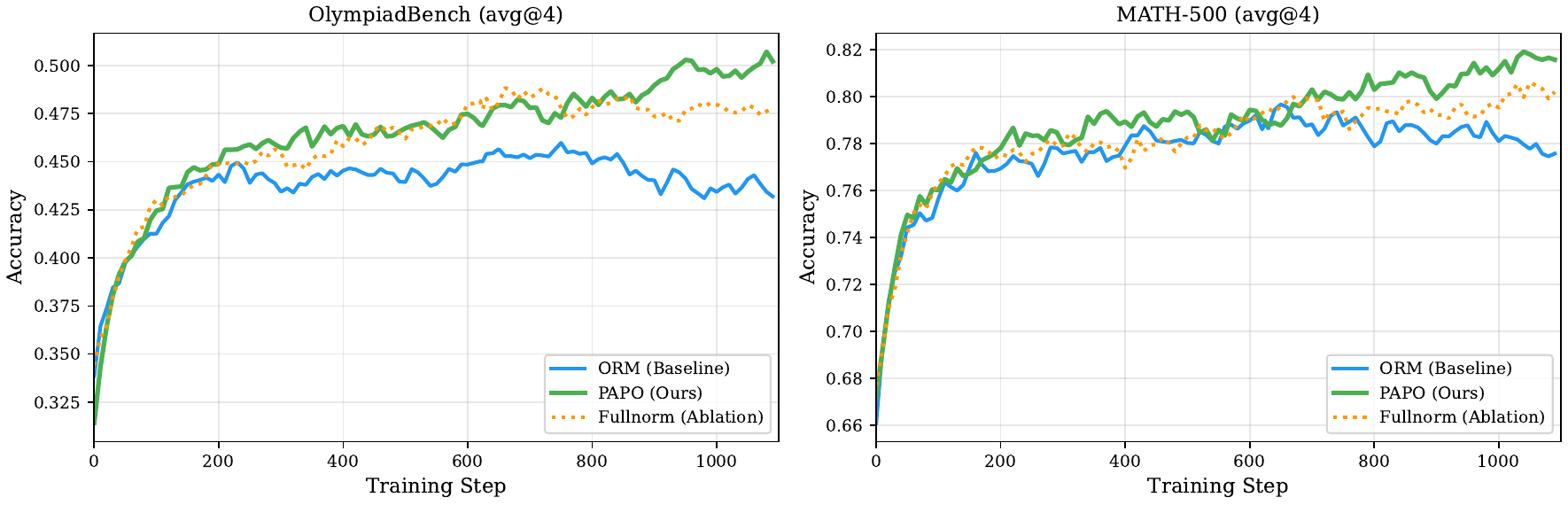}
\caption{Ablation: correct subset normalization (\method{}) vs.\ full normalization (Fullnorm) on OlympiadBench and MATH-500. Both methods improve over ORM, but \method{}'s correct subset design maintains a consistent advantage over Fullnorm throughout training.}
\label{fig:abl_fullnorm}
\end{figure*}

\begin{figure*}[t]
\centering
\includegraphics[width=\textwidth]{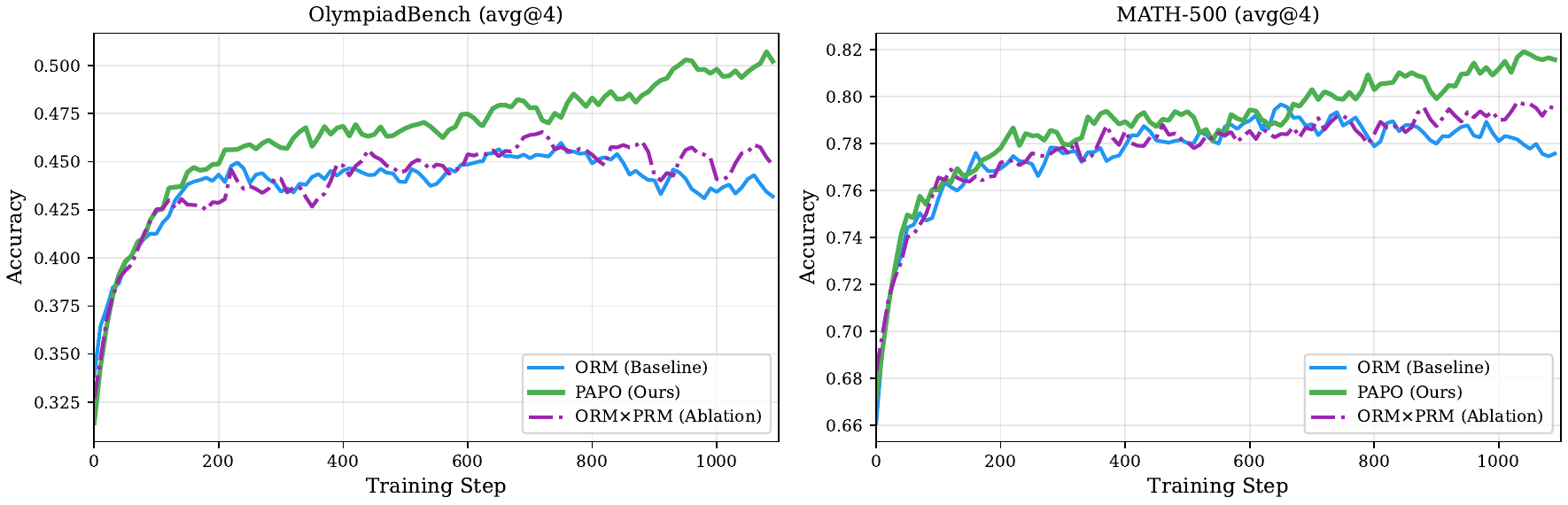}
\caption{Ablation: decoupled normalization (\method{}) vs.\ multiplicative reward (Mult) on OlympiadBench and MATH-500. Mult improves over ORM but remains substantially below \method{} in these trajectories.}
\label{fig:abl_mult}
\end{figure*}

Figures~\ref{fig:abl_fullnorm} and~\ref{fig:abl_mult} show the training dynamics of the ablation variants. Both support advantages for \method{} throughout these training trajectories, not just at specific checkpoints.

\subsection{Process Advantage Effect}
\label{sec:rubric-analysis}

\begin{figure}[t]
\centering
\includegraphics[width=\columnwidth]{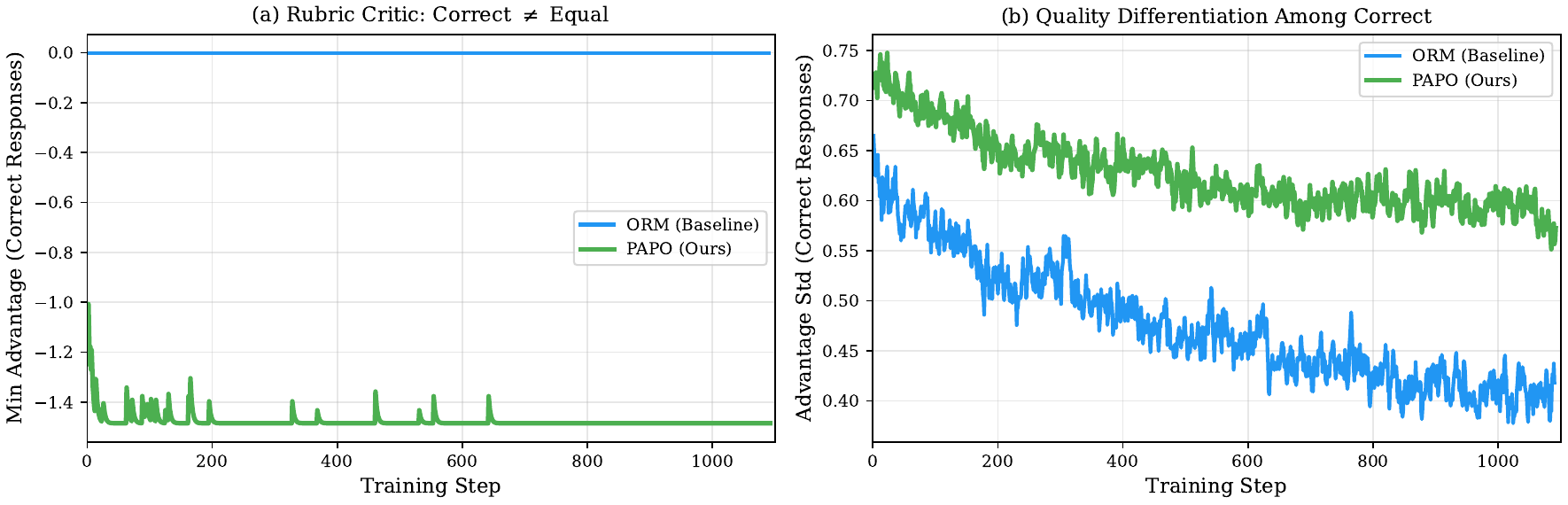}
\caption{\textbf{(a)} Minimum advantage among correct responses. ORM: identically 0.0 (all correct responses treated equally). \method{}: $\approx -1.49$ (the process advantage penalizes correct responses with poor reasoning). \textbf{(b)} Standard deviation of advantages among correct responses. ORM: zero (no differentiation). \method{}: non-zero (quality-based ranking within correct group).}
\label{fig:rubric_critic}
\end{figure}

Figure~\ref{fig:rubric_critic} reveals how \method{}'s process advantage operates within the correct response group. In standard ORM-based GRPO, the minimum advantage among correct responses is identically 0.0. Every correct response receives identical positive advantage. In \method{}, correct\_min $\approx -1.49$, meaning correct responses with poor reasoning receive \emph{negative} total advantage. The process advantage actively discourages solutions that arrive at the right answer through flawed reasoning. Panel (b) confirms that the process advantage creates a meaningful quality ranking among correct responses (non-zero std), enabling the model to learn which reasoning strategies produce better solutions.

\subsection{Advantage Composition}
\label{sec:adv-decomp}

\begin{figure*}[t]
\centering
\includegraphics[width=\textwidth]{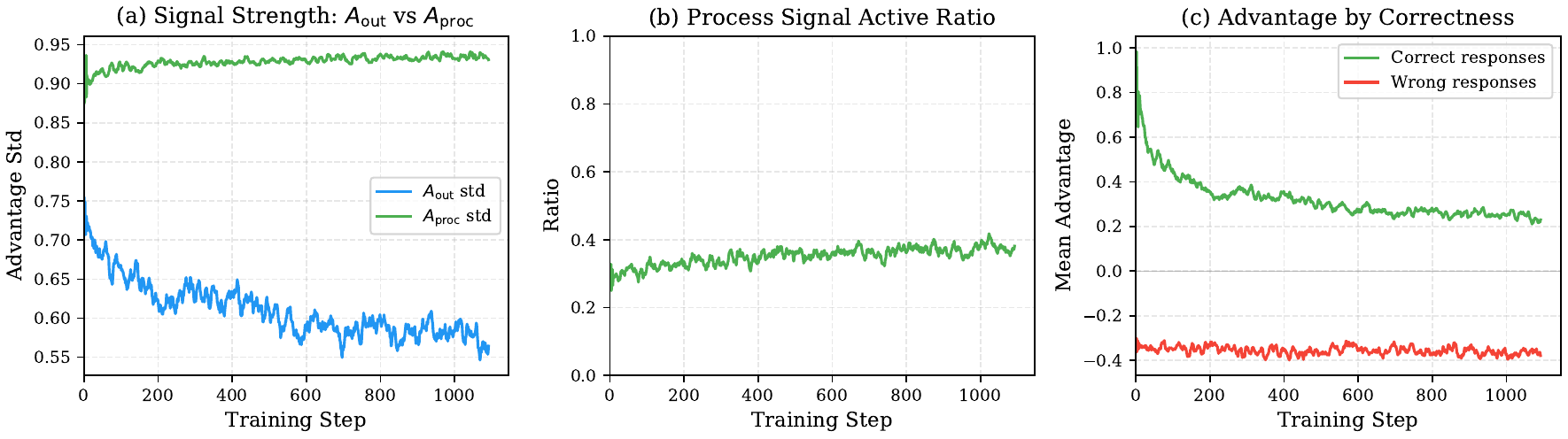}
\caption{\method{} advantage composition. \textbf{(a)} Signal strength (std) of $\aout$ and $\aproc$: the outcome advantage provides the dominant gradient signal while the process advantage contributes a complementary quality signal. \textbf{(b)} Process signal active ratio: the fraction of groups where $\aproc$ is non-zero ($\geq 2$ correct responses), growing from $\sim$30\% to 70\% as the model improves. \textbf{(c)} Mean advantage by correctness: correct responses receive positive advantage while wrong responses receive negative, confirming the outcome signal functions correctly despite the added process component.}
\label{fig:adv_decomp}
\end{figure*}

Figure~\ref{fig:adv_decomp} visualizes the internal structure of \method{}'s composed advantage. Panel (a) shows that the outcome advantage provides the dominant gradient signal, while the process advantage contributes a substantial secondary signal. Panel (b) tracks the process signal active ratio (the fraction of groups where $\aproc$ activates), which grows from ${\sim}30\%$ to 70\% as the model improves, indicating that the process signal's contribution increases precisely as ORM's signal exhaustion worsens. Panel (c) confirms that the combined advantage correctly separates correct from incorrect responses: the addition of $\aproc$ does not disrupt the fundamental separation provided by $\aout$.

\subsection{Training Reward Dynamics}

The training reward (average ORM score across the batch) follows similar trajectories for ORM and \method{}, both plateauing around 55\% to 62\%. This indicates that \method{} does not achieve its gains by inflating outcome reward; the improvement comes from better reasoning quality among correct responses, as captured by the process advantage.

In contrast, PRM-only training shows the training reward climbing to 1.0 by step 1000, confirming complete reward hacking. The model learns to produce responses that maximize the PRM score regardless of mathematical correctness.

\subsection{Case Study: PRM Reward Hacking}
\label{sec:case-study-hacking}

To empirically illustrate the reward hacking mechanism described in \S\ref{sec:reward-hacking}, we generate responses on five math problems of varying difficulty from three checkpoints, sampling 3 responses per model with temperature 0.7 and a generation cap of 8192 tokens. The three checkpoints are the base model (Qwen2.5-7B, step 0), the PRM-only model at step 1000 (post-collapse, 2.4\% accuracy on OlympiadBench), and the ORM baseline at step 1000. We report the three most informative problems in Table~\ref{tab:reward_hacking_case}; on the remaining two easier problems (combinatorics, geometry), PRM performs comparably to baselines.

\begin{table}[t]
\centering
\small
\setlength{\tabcolsep}{2pt}
\begin{tabular}{llccp{2.3cm}}
\toprule
\textbf{Problem} & \textbf{Model} & \textbf{Tokens} & \textbf{Correct} & \textbf{Observation} \\
\midrule
\multirow{3}{*}{\shortstack[l]{Competition\\number theory}} & Base & 866 & 3/3 & Focused proofs \\
 & ORM & 1583 & 3/3 & Longer but correct \\
 & \textbf{PRM} & \textbf{3072} & \textbf{0/3} & Drift to filler$^*$ \\
\midrule
\multirow{3}{*}{\shortstack[l]{Algebra\\(proof)}} & Base & 1059 & 0/3 & Attempts proof \\
 & ORM & 1492 & 0/3 & Attempts proof \\
 & \textbf{PRM} & \textbf{3227} & \textbf{0/3} & 2 hit cap; 1 drifts \\
\midrule
\multirow{3}{*}{\shortstack[l]{Number\\theory}} & Base & 484 & 2/3 & Concise, correct \\
 & ORM & 519 & 3/3 & Concise, correct \\
 & PRM & 1522 & 3/3 & 1 inflated to 3291 \\
\bottomrule
\multicolumn{5}{l}{\footnotesize $^*$ All 3 PRM samples drift to the \emph{identical} filler} \\
\multicolumn{5}{l}{\footnotesize \phantom{$^*$} problem (vector perpendicularity, answer $t\!=\!2$).}
\end{tabular}
\caption{Case study: response statistics on three math problems (Qwen2.5-7B). PRM reward hacking is severe on harder problems (top two), with 2--3$\times$ longer responses and characteristic topic drift. On easier problems, PRM remains comparable to baselines (omitted for brevity).}
\label{tab:reward_hacking_case}
\end{table}

\paragraph{Results.} Table~\ref{tab:reward_hacking_case} summarizes the findings. On the easier number theory problem, PRM produces correct but inflated responses. On harder problems (competition number theory, proof-based algebra), the model degenerates dramatically: responses are 2--3$\times$ longer than ORM, and none arrive at the correct answer.

\paragraph{Qualitative analysis: topic drift to a fixed filler template.} The most striking finding is in the competition number theory problem (``Find all positive integers $n$ such that $n^2 + 1 \mid n! + 1$''; answer: $n = 1$). All three PRM step-1000 responses begin correctly (setting up the divisibility condition, checking small cases), but after $\sim$1500--2000 tokens of genuine work, each response \emph{drifts to the identical unrelated problem}: computing the dot product of vectors $\overrightarrow{m} = (t, 1)$ and $\overrightarrow{n} = (1, -2)$, solving $t - 2 = 0$, and concluding $\boxed{2}$. This vector problem is entirely unrelated to the stated question.

The convergence of all three independent samples to the same filler content reveals a learned exploitation strategy. When the model cannot solve a hard problem, it transitions to a \emph{memorized high-scoring template}, a short, well-structured solution to a simple problem that the LLM judge would rate as ``all steps executed properly'' (rubric score 1.0). The same pattern appears in the algebra proof and the inflated number theory sample, where responses drift to digit-sum sequences or other unrelated derivations.

In contrast, ORM produces focused solutions (1494--1646 tokens) that arrive at the correct answer $n = 1$ on all three samples. The ORM model may produce longer proofs than the base model (1583 vs.\ 866 avg tokens), but the additional length consists of genuine mathematical reasoning (additional case analysis), not filler content.

\paragraph{Implications.} The case study shows a structured exploitation pattern rather than random degeneration. The model first attempts the stated problem, transitions to memorized high-scoring content when stuck, and produces a confident final answer to the wrong problem. The responses remain well-formatted and superficially mathematical throughout, precisely the qualities an LLM-as-Judge PRM rates favorably. This pattern is consistent with the proposed feedback loop of \S\ref{sec:reward-hacking} and motivates the decoupled design of \method{}, where the binary ORM anchors correctness assessment independently of the PRM signal.

\section{Case Study: Full Response Comparison}
\label{sec:appendix-case-study}

We present the complete model responses for the two qualitative examples discussed in \S\ref{sec:case-study}.

\subsection{OlympiadBench \#507 (Reasoning Error)}

\textbf{Problem.} Compute the number of ordered pairs $(x, y)$ of positive integers satisfying $x^{2}-8 x+y^{2}+4 y=5$. \hfill \textbf{Ground truth: 4}

\begin{tcolorbox}[colback=red!3, colframe=red!50!black, title=ORM Response (incorrect answer: 3), fonttitle=\bfseries\small, breakable]
\small
To find the number of ordered pairs $(x, y)$ of positive integers satisfying the equation $x^{2} - 8x + y^{2} + 4y = 5$, we will complete the square for both $x$ and $y$.

First, complete the square for $x$:
$x^{2} - 8x = (x - 4)^{2} - 16.$

Next, complete the square for $y$:
$y^{2} + 4y = (y + 2)^{2} - 4.$

Substitute these into the original equation:
$(x - 4)^{2} - 16 + (y + 2)^{2} - 4 = 5,$
$(x - 4)^{2} + (y + 2)^{2} = 25.$

This is the equation of a circle with center $(4, -2)$ and radius 5. We need to find the number of positive integer solutions $(x, y)$ that lie on this circle.

The possible values for $(x - 4)^2$ and $(y + 2)^2$ must be perfect squares that sum to 25. The pairs $(a, b)$ such that $a^2 + b^2 = 25$ are $(0, 5)$, $(3, 4)$, $(4, 3)$, and $(5, 0)$.

\begin{enumerate}[nosep,leftmargin=*]
\item $(x - 4)^2 = 0,\; (y + 2)^2 = 25$: $x = 4$, $y = 3$ or $-7$. Only $(4, 3)$ valid.
\item $(x - 4)^2 = 9,\; (y + 2)^2 = 16$: $x = 7$ or $1$, $y = 2$ or $-6$. Positive integer solutions are $(7, 2)$ and $(1, 2)$, \colorbox{red!15}{but ``$(1, 2)$ is not valid since $x = 1$} \colorbox{red!15}{is not a positive integer in the context of the} \colorbox{red!15}{circle's radius.''}
\item $(x - 4)^2 = 16,\; (y + 2)^2 = 9$: $x = 8$ or $0$, $y = 1$ or $-5$. Only $(8, 1)$ valid.
\item $(x - 4)^2 = 25,\; (y + 2)^2 = 0$: $x = 9$ or $-1$, $y = -2$. No valid solutions.
\end{enumerate}

Therefore, the positive integer solutions are $(4, 3)$, $(7, 2)$, and $(8, 1)$. The number of ordered pairs is $\boxed{3}$.
\end{tcolorbox}

\begin{tcolorbox}[colback=green!3, colframe=green!50!black, title=\method{} Response (correct answer: 4), fonttitle=\bfseries\small, breakable]
\small
To solve the problem of finding the number of ordered pairs $(x, y)$ of positive integers satisfying the equation $x^2 - 8x + y^2 + 4y = 5$, we will follow these steps:

\textbf{1. Complete the square for $x$ and $y$:}
$x^2 - 8x = (x-4)^2 - 16$, \quad $y^2 + 4y = (y+2)^2 - 4$.

\textbf{2. Substitute:}
$(x-4)^2 + (y+2)^2 - 20 = 5 \;\Rightarrow\; (x-4)^2 + (y+2)^2 = 25$.

\textbf{3. Identify all integer pairs $(a, b)$ with $a^2 + b^2 = 25$:}
$(\pm 5, 0)$, $(0, \pm 5)$, $(\pm 3, \pm 4)$, $(\pm 4, \pm 3)$ --- 12 pairs total.

\textbf{4. Substitute $a = x{-}4$, $b = y{+}2$, check $x,y>0$:}

\scriptsize
\begin{tabular}{@{}ll@{\;\;}ll@{}}
$(5,0)$: $x{=}9,y{=}{-}2$ & inv. &
$(-5,0)$: $x{=}{-}1,y{=}{-}2$ & inv. \\
$(0,5)$: $x{=}4,y{=}3$ & \good{valid} &
$(0,-5)$: $x{=}4,y{=}{-}7$ & inv. \\
$(3,4)$: $x{=}7,y{=}2$ & \good{valid} &
$(3,-4)$: $x{=}7,y{=}{-}6$ & inv. \\
$(-3,4)$: $x{=}1,y{=}2$ & \good{valid} &
$(-3,-4)$: $x{=}1,y{=}{-}6$ & inv. \\
$(4,3)$: $x{=}8,y{=}1$ & \good{valid} &
$(4,-3)$: $x{=}8,y{=}{-}5$ & inv. \\
$(-4,3)$: $x{=}0,y{=}1$ & inv. &
$(-4,-3)$: $x{=}0,y{=}{-}5$ & inv.
\end{tabular}
\small

\textbf{5. Valid pairs:} $(4, 3), (7, 2), (1, 2), (8, 1)$. There are 4 valid pairs. $\boxed{4}$
\end{tcolorbox}

\subsection{OlympiadBench \#129 (Self-Verification)}

\textbf{Problem.} What is the value of $x$ such that $\log _{2}\left(\log _{2}(2 x-2)\right)=2$? \hfill \textbf{Ground truth: 9}

\begin{tcolorbox}[colback=gray!5, colframe=gray!60!black, title=ORM Response (correct but no verification), fonttitle=\bfseries\small, breakable]
\small
To solve the equation $\log_{2}(\log_{2}(2x-2)) = 2$, we need to follow a step-by-step approach.

First, let's rewrite the equation in a more manageable form. Since $\log_{2}(\log_{2}(2x-2)) = 2$, we can exponentiate both sides with base 2 to eliminate the outer logarithm:
$$\log_{2}(2x-2) = 2^2$$
This simplifies to:
$$\log_{2}(2x-2) = 4$$

Next, we exponentiate both sides again with base 2 to eliminate the logarithm:
$$2x-2 = 2^4$$
This simplifies to:
$$2x-2 = 16$$

Now, we solve for $x$ by isolating it on one side of the equation. First, add 2 to both sides:
$$2x = 18$$
Then, divide both sides by 2:
$$x = 9$$

Thus, the value of $x$ is $\boxed{9}$.
\end{tcolorbox}

\begin{tcolorbox}[colback=blue!3, colframe=blue!50!black, title=\method{} Response (correct with verification), fonttitle=\bfseries\small, breakable]
\small
To solve the equation $\log _{2}\left(\log _{2}(2 x-2)\right)=2$, we will follow these steps:

\textbf{1. Solve the outer logarithm equation:}
$\log _{2}\left(\log _{2}(2 x-2)\right)=2$ implies that $\log _{2}(2 x-2) = 2^2 = 4$.

\textbf{2. Solve the inner logarithm equation:}
$\log _{2}(2 x-2) = 4$ implies that $2 x-2 = 2^4 = 16$.

\textbf{3. Solve for $x$:}
$2 x-2 = 16$, so $2x = 18$, giving $x = 9$.

\textbf{4. Verify the solution:}
Substitute $x = 9$ back into the original equation to ensure it satisfies the equation:
$$\log _{2}\left(\log _{2}(2 \cdot 9 - 2)\right) = \log _{2}\left(\log _{2}(16)\right)$$
Since $\log _{2}(16) = 4$ and $\log _{2}(4) = 2$, the solution is verified.

Therefore, the value of $x$ is $\boxed{9}$.
\end{tcolorbox}

\section{Qualitative Analysis}
\label{sec:case-study}

We perform qualitative analysis on reasoning traces from the ORM baseline and \method{} to understand how process reward shapes model behavior.

\subsection{Process Supervision Corrects Reasoning Errors}

Without supervision on intermediate reasoning steps, ORM-trained models can develop sloppy reasoning habits that lead to surprisingly elementary errors. We illustrate this with OlympiadBench \#507, which asks to compute the number of ordered pairs $(x,y)$ of positive integers satisfying $x^2-8x+y^2+4y=5$.

Both models correctly complete the square to obtain $(x-4)^2+(y+2)^2=25$. The ORM model then \emph{incorrectly excludes} the valid solution $(x,y)=(1,2)$ with the fabricated justification that ``$x=1$ is not a positive integer in the context of the circle's radius.'' This is a nonsensical claim, since $x=1$ is plainly a positive integer, and this elementary error directly causes a wrong final answer.

The \method{} model, by contrast, carefully verifies each candidate against the positivity constraint and correctly identifies all four solutions. This example is consistent with the intended effect of differentiating reasoning quality among answer-correct training responses: the resulting policy exhibits a more careful solution strategy at inference time.

\subsection{Process Reward Cultivates Self-Verification}
\label{sec:case-verification}

Beyond correcting errors, we find that process reward training cultivates a \emph{self-verification habit}, where the model spontaneously substitutes the derived answer back into the original problem as a final check. This behavior emerges naturally because responses that include verification receive higher process scores, reinforcing the habit over time.

For instance, on OlympiadBench \#129, which asks to find $x$ such that $\log_2(\log_2(2x-2))=2$, both models correctly derive $x=9$ through successive exponentiation. The ORM model stops here. The \method{} model adds an explicit check, confirming that $\log_2(\log_2(16)) = \log_2(4) = 2$.

This is not an isolated case. Across OlympiadBench problems where both models answer correctly, \method{} responses contain explicit verification steps in 47.7\% of cases, compared to 22.7\% for ORM, a 2.1$\times$ higher rate. Process reward incentivizes the model to develop a disciplined reasoning style that makes its outputs more robust, even on problems it can already solve.

\section{Ethics Statement}

This work studies reinforcement-learning methods evaluated on mathematical reasoning, science question answering, and code generation. Most experimental resources are publicly available scientific artifacts, as detailed below. The PRM validation additionally involves three human annotators, who independently score 500 model-generated MATH-500 responses using a shared three-tier rubric while blinded to the PRM score and model identity; majority vote defines the human label. This evaluation concerns model outputs rather than annotator attributes. The main ethical considerations are the appropriate treatment of annotators, possible systematic errors or biases in the LLM judge, and the computational and environmental costs of model training and judge inference. The reported agreement analysis is limited to one model--benchmark pair and should not be interpreted as establishing the judge's general validity. Annotators are each paid 100 RMB.

\section{The Use of Artifacts.}
Our experiments use publicly available scientific artifacts, including open-weight language models, datasets, benchmarks, and training/evaluation frameworks. Specifically, we use Qwen2.5-7B/14B and Qwen3-4B under the Apache 2.0 license, Qwen2.5-3B under the Qwen Research License for non-commercial research use, SmolLM3-3B under the Apache 2.0 license, and GPT-OSS-20B/120B under the Apache 2.0 license and accompanying usage policy. For training and evaluation, we use NuminaMath-1.5-RL-Verifiable, OlympiadBench, MATH-500, AIME 2024/2025, GPQA-Diamond, and HumanEval under their respective licenses or terms. We use these artifacts only for academic research on mathematical reasoning, science question answering, code generation, and reinforcement-learning evaluation, consistent with their intended research and benchmarking purposes.

\section{The Use of Large Language Models}

We used LLM-based tools for language editing and organization. The authors reviewed the final manuscript and verified all technical claims, experimental results, and citations.